\documentclass[11pt]{article}

\usepackage[preprint]{acl}

\usepackage{times}
\usepackage{latexsym}

\usepackage[T1]{fontenc}

\usepackage[utf8]{inputenc}

\usepackage{microtype}

\usepackage{inconsolata}

\usepackage{graphicx}
\usepackage{booktabs}
\usepackage{tabularx}
\usepackage{multirow}
\usepackage{amsmath}
\usepackage{amssymb}
\usepackage{xcolor}

\newcommand{\sse}{S$^3$E}

\newcommand{\sStressBand}{\ensuremath{\overline{S}^{\mathrm{stress}}_{\mathrm{top}}}}

\newcommand{\CA}{\textsc{CA}}
\newcommand{\SO}{\textsc{SO}}
\newcommand{\SI}{\textsc{SI}}
\newcommand{\BE}{\textsc{BE}}
\newcommand{\dpres}{\ensuremath{\Delta^{\mathrm{pres}}}}
\newcommand{\dlex}{\ensuremath{\Delta^{\mathrm{lex}}}}
\newcommand{\drand}{\ensuremath{\Delta^{\mathrm{rand}}}}
\newcommand{\dE}{\ensuremath{\Delta_{\mathrm{E5}}}}

\newcolumntype{Y}{>{\RaggedRight\arraybackslash}X}

\title{When Correct Decisions Hide Internal Stress: Decision-State Probing in Multimodal Language Models}

\author{
  \textbf{Haoran Zhao\textsuperscript{1}},
  \textbf{Soyeon Caren Han\textsuperscript{1}\thanks{Corresponding author.}},
  \textbf{Eduard Hovy\textsuperscript{1}}
\\
  \textsuperscript{1}The University of Melbourne
\\
  {\normalsize
    \textsuperscript{1}\texttt{haoran.zhao.2@student.unimelb.edu.au},
    \textsuperscript{*}\texttt{Caren.Han@unimelb.edu.au},
    \textsuperscript{1}\texttt{eduard.hovy@unimelb.edu.au}
  }
}

\begin{document}
\maketitle

\begin{abstract}
Multimodal language models are typically evaluated through external behavior: selecting the correct image--text match, rejecting unsupported captions, or answering visual queries correctly. However, correct behavior alone does not show that the model's internal decision state remains stable under controlled semantic stress. We study this gap through \sse{} (Structured Semantic Stress Evaluation), a framework for analyzing behavior--internal decoupling in multimodal language models. \sse{} uses a positive-anchored A/B forced-choice setup in which an image-supported caption is contrasted against semantic stress candidates under both original and swapped option orders, while hidden states are extracted at the pre-answer decision state. We focus on \emph{strict-correct} trials, where the model consistently selects the correct caption across both orders. Rather than treating arbitrary hidden-state variation as evidence of instability, we measure whether semantic-conflict candidates induce excess decision-state displacement relative to meaning-preserving controls. Across Qwen3VL, Gemma3, and InternVL3, semantic stress consistently produces positive selected-layer excess displacement over lexical controls despite correct forced-choice behavior, while comparisons against random negatives are model-dependent. We interpret this as a scoped decision-state stress-sensitivity signal rather than evidence of downstream failure or hallucination. Our results suggest that forced-choice correctness alone is not a sufficient certificate of invariant internal decision geometry.
\end{abstract}

\begin{figure*}[t]
\centering
\includegraphics[width=\textwidth]{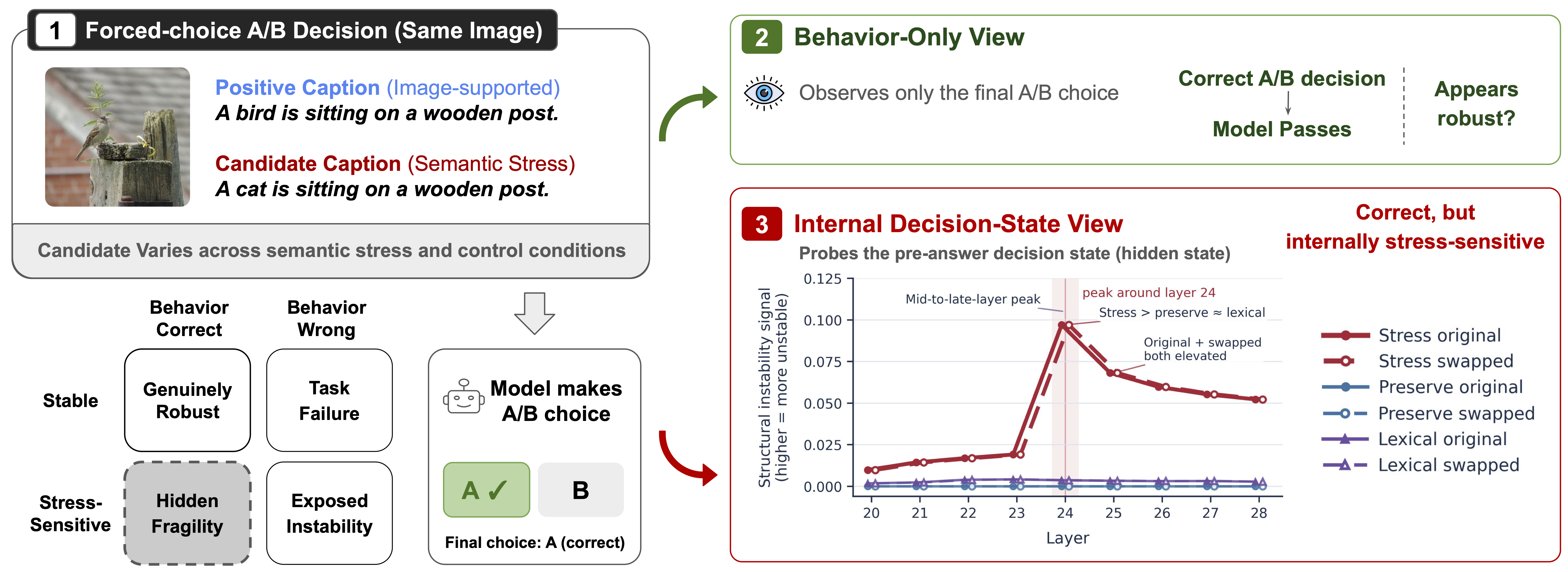}
\caption{Motivating example of behavior--internal decoupling. Under the behavior-only view, a model appears superficially robust when it makes the correct A/B decision. \sse{} additionally probes the pre-answer decision state and can reveal a hidden stress-sensitivity signal: a strict-correct decision may still exhibit elevated preserve-anchored decision-state displacement under semantic stress.}
\label{fig:teaser}
\end{figure*}

\section{Introduction}

Multimodal language models (MLLMs) are commonly evaluated through externally observable behavior. A model is treated as reliable if it selects the correct caption, rejects an unsupported description, or answers an image-grounded query correctly. This behavior-centric paradigm has been essential for measuring progress in multimodal understanding, hallucination detection, and compositional reasoning \citep{rohrbach2018object,thrush2022winoground,yuksekgonul2023aro,hsieh2023sugarcrepe,li2023pope,guan2024hallusionbench}. 
Yet correct behavior alone does not guarantee that the model's internal decision state remains stable under semantic stress.
This paper studies a specific evaluation blind spot: a model may be \emph{correct but internally stress-sensitive}. Under structured semantic stress, the model can make the same correct forced-choice decision while its pre-answer decision state exhibits elevated displacement relative to semantic-preserving controls. In such cases, behavior-only metrics report success, but they miss an internal decision-state stress-sensitivity signal. Figure~\ref{fig:teaser} illustrates this gap between behavior-only evaluation and the internal-state view.
We focus on \emph{structured semantic stress}: carefully controlled meaning-changing perturbations to image--caption pairs. Rather than arbitrary corruptions, these perturbations act as interpretable probes of how the model's pre-answer decision geometry changes when a candidate caption introduces semantic conflict with the image-supported description.


We introduce \sse{} (Structured Semantic Stress Evaluation), an evaluation lens for systematically measuring behavior--internal decoupling in MLLMs. This paper uses a \emph{positive-anchored A/B forced-choice interface}. Each trial presents an image, an image-supported positive caption, and a candidate caption. The model must choose the caption that better matches the image, with both the original and swapped A/B orders evaluated. We analyze hidden states at the \emph{pre-answer decision state}, immediately before the model emits the forced-choice answer.
Crucially, we do not interpret arbitrary variation in the hidden state as evidence of stress sensitivity: different captions should naturally induce different representations. Instead, under matched correct behavior, we ask whether semantic-conflict candidates produce excess pre-answer decision-state displacement relative to meaning-preserving lexical controls.



To operationalize this distinction, our evaluation proceeds in four stages. First, we make the base--preserve--stress substrate self-contained in the present manuscript rather than asking the reader to rely on a concurrent anonymous benchmark submission. Second, we verify that the forced-choice interface makes semantic stress behaviorally nontrivial. Third, we condition on \emph{strict-correct} semantic-stress decisions, where the model selects the positive caption in both A/B orders against the stress candidate. Fourth, we report selected-layer specificity contrasts showing whether semantic stress exceeds both preserve and lexical controls, while treating random-negative contrasts and limited perturbation checks as separate generic-mismatch sanity checks rather than primary evidence.
Our contributions are:
\begin{itemize}
    \item We formalize \emph{pre-answer decision-state stress sensitivity under structured semantic stress} as a scoped diagnostic target for MLLMs.
    \item We propose a positive-anchored A/B forced-choice protocol, comparing external correctness with pre-answer decision-state diagnostics under original and swapped option orders.
    \item We instantiate the protocol on a public-source-derived base--preserve--stress substrate and provide the construction roles, curation summary, source composition, metadata, and validation checks needed to make the present analysis self-contained.
    \item We introduce a strict-correct, control-normalized hidden-state analysis for behavior-conditioned decision states. Conditioning on trials that are correct under both option orders, we measure selected-layer stress-minus-lexical excess displacement, keep preserve/lexical controls separate from random mismatch references, and find positive stress-minus-lexical effects across high-support runs while stress-minus-random remains model-dependent.
\end{itemize}

\section{Related Work}

\paragraph{Output-based evaluation of MLLMs.}
Multimodal model evaluation is largely organized around externally observable behavior, including caption quality, VQA accuracy, multiple-choice correctness, image--text matching, and hallucination or grounding judgments. Broad suites such as MME, MMBench, and MM-Vet evaluate perception, reasoning, OCR, knowledge, spatial understanding, and language generation \citep{fu2023mme,liu2024mmbench,yu2024mmvet}, while hallucination and grounding benchmarks assess whether model outputs are supported by the image \citep{rohrbach2018object,li2023pope,guan2024hallusionbench}. These benchmarks are essential for measuring external reliability; \sse{} complements them by conditioning on correct forced-choice behavior and measuring how the pre-answer decision state responds to controlled semantic stress.

\paragraph{Compositional, contrastive, and metamorphic stress tests.}
Controlled image--text contrasts have been widely used to probe compositionality and fine-grained semantic grounding. Winoground uses minimally contrastive image--caption pairs \citep{thrush2022winoground}; ARO targets attribution, relation, and order sensitivity \citep{yuksekgonul2023aro}; and SugarCrepe improves hard-negative construction for image--text compositionality \citep{hsieh2023sugarcrepe}. This line of work shares the idea that controlled semantic variation can expose failures hidden by aggregate accuracy. Our experiments use a structured base--preserve--stress substrate from a concurrent anonymized benchmark submission~\citep{anonymous2026semanticstress}; that work introduces the resource and black-box behavioral metrics, while the present paper uses for targeted white-box pre-answer decision-state analysis.

\paragraph{Representation diagnostics and probing.}
Probing and representation-analysis work has shown that hidden states can encode structure not directly visible from outputs \citep{hewitt2019structural,rogers2020primer}. Related methods compare representations across layers or models using SVCCA and CKA \citep{raghu2017svcca,kornblith2019similarity}, and studies of contextual embeddings document anisotropy and concentration effects in transformer spaces \citep{ethayarajh2019contextual}. Recent multimodal explanation work further analyzes internal token interactions and attribution fidelity in MLLMs \citep{liang2025intramodal}. \sse{} follows this representation-diagnostic perspective but uses hidden-state geometry as an evaluation lens: it measures control-normalized pre-answer displacement under structured semantic stress after the model has already made the correct forced-choice decision.

\paragraph{Interventions and positioning.}
Mechanistic and causal interpretability methods test whether internal components are functionally implicated in behavior, for example through causal mediation analysis, interchange interventions, causal tracing, and activation steering \citep{vig2020causal,geiger2021causal,meng2022locating,rimsky2024steering}. We use this perspective only as supplementary context for limited perturbation checks; the main contribution is not a causal mechanism claim, but a scoped diagnostic lens connecting behavior-centric multimodal evaluation with representation-level measurements on strict-correct decisions.

\section{Evaluation Target}
\label{sec:target}

\subsection{Structured Semantic Stress}
Given an image $I$ and a caption $c^+$ supported by the image, a semantic-stress candidate $c^{\mathrm{stress}}$ changes a targeted semantic factor while preserving enough linguistic form to make the contrast interpretable. We consider object substitutions, attribute substitutions, relation changes, and compositional/order confusions. 

\paragraph{Substrate and contribution boundary.}
The candidate set used here is drawn from a concurrent anonymized benchmark submission on structured base--preserve--stress evaluation~\citep{anonymous2026semanticstress}. That benchmark paper defines each source image--caption example through a base caption, a meaning-preserving preserve caption, and a meaning-changing stress caption. The present paper uses that controlled substrate for a different purpose: white-box decision-state analysis. We therefore do not treat benchmark construction or black-box model ranking as contributions of this paper. Conversely, the current manuscript includes the substrate details necessary to interpret the hidden-state analysis without relying on the concurrent submission being accepted.

\paragraph{Public sources and curation.}
The substrate is derived from public compositional image--text sources: Winoground~\citep{thrush2022winoground}, SugarCrepe~\citep{hsieh2023sugarcrepe}, and ARO~\citep{yuksekgonul2023aro}.
The hidden-state experiments in this paper use the released benchmark core, which contains 11,522 stress variants over 7,838 images. Curation applies validity filtering, rule-based severity scoring, duplicate reduction, per-image caps, source balancing, and slice-aware sampling. Additional pool statistics, including the broader candidate snapshot, the feasible three-source pool, and the matched analysis subset used by the concurrent benchmark, are reported in Appendix~\ref{app:benchmark_substrate}.


\begin{table}[t]
\centering
\scriptsize
\setlength{\tabcolsep}{3.0pt}
\begin{tabularx}{\columnwidth}{lX}
\toprule
\textbf{Substrate field} & \textbf{Summary used for this paper} \\
\midrule
Source pools & Winoground, SugarCrepe, and ARO public image--caption benchmarks. \\
Candidate pool & 426,278 stress variants; feasible pool: 418,969 variants over 30,531 images. \\
Released views & Benchmark core used here: 11,522 variants over 7,838 images; concurrent behavioral matched subset: 4,000 variants over 3,401 images. \\
Curation & Source/readiness filtering, severity scoring, duplicate reduction, per-image caps, source balancing, and slice-aware sampling. \\
Metadata & Stress type, operation metadata, rule-based severity level, source family, and diagnostic slice. \\
Human audit & 120 stratified examples with five annotators per item; preserve validity 96.7\%, stress validity 95.0\%, stress naturalness 89.2\%, and single-factor control 88.3\%. \\
\bottomrule
\end{tabularx}
\caption{Self-contained summary of the public-source-derived base--preserve--stress substrate used by \sse{}. The concurrent benchmark submission provides the full resource; the present paper uses the 11,522-variant benchmark core for hidden-state evaluation.}
\label{tab:substrate_summary}
\end{table}

\begin{table}[t]
\centering
\scriptsize
\setlength{\tabcolsep}{3.0pt}
\begin{tabularx}{\columnwidth}{lXX}
\toprule
\textbf{Source} & \textbf{Public role} & \textbf{Use in \sse{}} \\
\midrule
Winoground & Compositional image--caption pairs & Compositional/order semantic stress candidates and high-overlap contrastive cases. \\
SugarCrepe & Hard image--text negatives & Object, attribute, and relation-style stress candidates. \\
ARO & Attribution, relation, and order probes & Attribute, relation, and order stress candidates. \\
\bottomrule
\end{tabularx}
\caption{Public source benchmarks underlying the anonymized evaluation subset. The current paper reports the source roles needed for the hidden-state analysis; full resource construction is part of the concurrent anonymous benchmark submission.}
\vspace{-1em}
\label{tab:source_substrates}
\end{table}

\paragraph{Condition roles in \sse{}.}
The base--preserve--stress benchmark roles are mapped into the condition roles needed for the present internal analysis. We write $c^+$ for the image-supported positive caption used as the A/B gold target and as the geometric reference state. The preserve control $c^{\mathrm{pres}}$ is a same-image meaning-preserving rewrite that should remain semantically correct; it tests whether ordinary semantic-preserving reformulation stays internally stable. The lexical control $c^{\mathrm{lex}}$ is a distinct meaning-preserving surface-control candidate that estimates the effect of word-level and caption-surface variation. The stress caption is the targeted meaning-changing candidate, and a random negative $c^{\mathrm{rand}}$ sampled outside the source item is used as a generic mismatch reference. 
\begin{table}[t]
\centering
\scriptsize
\setlength{\tabcolsep}{3.0pt}
\begin{tabularx}{\columnwidth}{lXX}
\toprule
\textbf{Role} & \textbf{Construction} & \textbf{Use in this paper} \\
\midrule
Positive $c^+$ & Image-supported caption & Gold target for stress/random A/B behavior and geometric reference state $h_+$ in Eq.~\ref{eq:primary_score}. \\
Preserve $c^{\mathrm{pres}}$ & Meaning-preserving rewrite or equivalent expression for the same image & Semantic-preserving control: tests whether a correct reformulation remains internally stable. \\
Lexical $c^{\mathrm{lex}}$ & Meaning-preserving surface-control caption distinct from $c^{\mathrm{pres}}$ & Lexical/surface-form control: estimates ordinary wording and caption-distance effects; not used to define behavioral correctness. \\
Stress $c^{\mathrm{stress}}$ & Meaning-changing object, attribute, relation, or compositional foil & Target semantic-stress condition. \\
Random $c^{\mathrm{rand}}$ & Caption from another source item & Generic mismatch reference, not the primary semantic baseline. \\
\bottomrule
\end{tabularx}
\caption{Condition roles in this paper. The concurrent benchmark defines the base--preserve--stress resource; \sse{} uses the positive caption as the forced-choice gold target and geometric reference, while preserve and lexical are distinct meaning-preserving controls for control-normalized hidden-state analysis.}
\label{tab:condition_roles}
\end{table}
Table~\ref{tab:condition_roles} summarizes the condition roles used in this paper. The positive caption serves as the geometric reference. Preserve and lexical controls are kept separate: preserve anchors the image-supported meaning, while the lexical condition captures meaning-preserving caption-surface variation. Random negatives are reported separately as a generic mismatch reference.

\paragraph{Control normalization.}
To isolate semantic-specific effects beyond caption variation, we use a control-normalized, positive-anchored cosine-distance diagnostic. We ask whether semantic-conflict candidates produce excess preserve-anchored decision-state displacement beyond the lexical control under matched correct behavior.

\subsection{Positive-Anchored Forced Choice}

Each evaluation item contains an image $I$, a positive caption $c^+$, and a candidate caption $c$. The model is prompted to choose which of two caption options better matches the image. Both option orders are systematically evaluated:
\begin{equation}
\begin{aligned}
o=\mathrm{orig}:&\ [c^+, c],\\
o=\mathrm{swap}:&\ [c, c^+].
\end{aligned}
\label{eq:orders}
\end{equation}
For negative candidates such as stress and random captions, the correct behavior is to select $c^+$. We define a source item as \emph{strict-correct} for the main analysis when the model selects $c^+$ in both option orders on the semantic-stress trial. This removes simple A/B position bias and ensures that the internal analysis is conditioned on stable external behavior under the target stress condition. Preserve and lexical controls are meaning-preserving alternatives, so choosing $c^+$ over those controls is not treated as a behavioral correctness requirement; their states are retained as internal controls for the same stress-strict source items.
This interface is distinct from free-form caption generation, broad VQA, and yes/no verification. It is a caption-level, positive-anchored, binary forced-choice semantic verification interface.

\subsection{Pre-Answer Decision State}

We extract hidden states at the pre-answer decision state, immediately before the model emits the forced-choice answer token. This choice matters: the analysis is not about arbitrary token-level differences between two captions, but about the internal state at the moment of semantic decision. Let $h_{i,o,k}^{(m,\ell)}$ denote the hidden state for model $m$, layer $\ell$, source item $i$, option order $o$, and condition $k$.

\begin{figure*}[t]
\centering
\includegraphics[width=\textwidth]{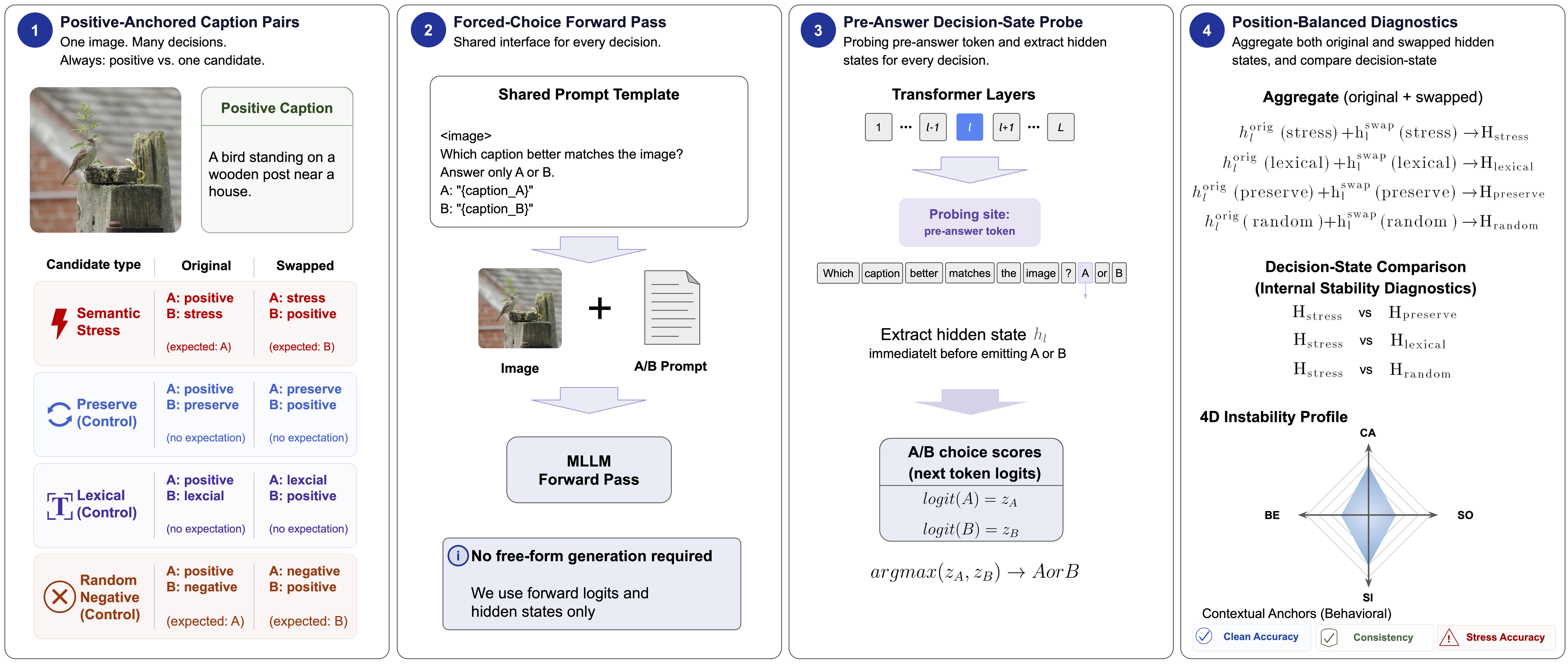}
\caption{Overview of the \sse{} pipeline. Each item is converted into positive-anchored caption pairs, evaluated through a shared A/B forced-choice pass, probed at the pre-answer state, and aggregated across swapped orders. The resulting position-balanced diagnostics use the positive caption as the reference state, compare stress against preserve, lexical, and random controls, and summarize diagnostics through the \CA{}, \SO{}, \SI{}, and \BE{} profile.}
\label{fig:pipeline}
\end{figure*}

Figure~\ref{fig:pipeline} summarizes the full evaluation pipeline, from positive-anchored candidate construction to position-balanced diagnostic aggregation.

\subsection{Primary Score and Control-Normalized Stress Sensitivity}
\label{sec:primary_score}

The diagnostic target is not raw hidden-state movement: different captions should naturally induce different pre-answer representations. We therefore use a \emph{positive-anchored pre-answer cosine distance}. For model $m$, layer $\ell$, source item $i$, option order $o$, and candidate condition $k\in\{\mathrm{pres},\mathrm{lex},\mathrm{stress},\mathrm{random}\}$, let $h_{i,o,k}^{(m,\ell)}$ denote the language-side hidden state at the pre-answer decision token. We define
\begin{equation}
S_{i,o,k}^{(m,\ell)}
= 1 - \cos\!\left(h_{i,o,k}^{(m,\ell)}, h_{i,o,+}^{(m,\ell)}\right),
\label{eq:primary_score}
\end{equation}
where $h_{i,o,+}^{(m,\ell)}$ is the matched positive-caption reference state for the same image, source item, option order, model, and layer. Both vectors are normalized through the cosine operation; we do not pool generated text, use free-form decoding states, center or whiten activations, or apply dimensionality reduction before computing this score. 
On source items that are strict-correct on semantic-stress trials, we compute preserve and lexical excess contrasts after averaging the two A/B option orders:
\begin{equation}
\Delta^{\mathrm{pres}}_{i}
= \frac{1}{2}\sum_{o\in\{\mathrm{orig},\mathrm{swap}\}}
\left[ S_{i,o,\mathrm{stress}} - S_{i,o,\mathrm{pres}} \right],
\label{eq:dpres}
\end{equation}
\begin{equation}
\Delta^{\mathrm{lex}}_{i}
= \frac{1}{2}\sum_{o\in\{\mathrm{orig},\mathrm{swap}\}}
\left[ S_{i,o,\mathrm{stress}} - S_{i,o,\mathrm{lex}} \right].
\label{eq:dlex}
\end{equation}
Here $\mathrm{pres}$ denotes the meaning-preserving control and $\mathrm{lex}$ denotes the lexical surface-form control for the same source item. \dpres{} asks whether structured semantic stress induces more displacement than a meaning-preserving reformulation, while \dlex{} asks whether the effect exceeds ordinary wording and caption-surface variation. The headline semantic-specificity result uses \dlex{} as the primary paired contrast.
For context, we also compute a generic mismatch contrast:
\begin{equation}
\Delta^{\mathrm{rand}}_{i}
= \frac{1}{2}\sum_{o\in\{\mathrm{orig},\mathrm{swap}\}}
\left[ S_{i,o,\mathrm{stress}} - S_{i,o,\mathrm{random}} \right].
\label{eq:drand}
\end{equation}
Here $\mathrm{random}$ denotes a caption sampled outside the source item. 
We additionally report \drand{} as a generic mismatch reference. The primary semantically specific baseline remains lexical control.



\begin{table}[t]
\centering
\small
\setlength{\tabcolsep}{4pt}
\renewcommand{\arraystretch}{1.15}
\begin{tabularx}{\columnwidth}{@{}>{\bfseries}l
  >{\raggedright\arraybackslash}X
  >{\raggedright\arraybackslash}X@{}}
\toprule
&
\shortstack[l]{\textbf{Decision-state}\\\textbf{stable}}
&
\shortstack[l]{\textbf{Decision-state}\\\textbf{stress-sensitive}} \\
\midrule
Strict-correct
& Robust decision
& Hidden stress-sensitivity signal \\
\addlinespace[2pt]
Incorrect
& Exposed behavior failure
& Exposed stress sensitivity \\
\bottomrule
\end{tabularx}
\caption{Behavior--internal decoupling in the A/B forced-choice interface. Behavior-only evaluation observes whether the semantic-stress decision is correct; \sse{} asks whether stress-trial strict-correct decisions remain decision-state stable under structured semantic stress.}
\label{tab:decoupling_quadrants}
\end{table}

Table~\ref{tab:decoupling_quadrants} summarizes the evaluation target. \sse{} focuses on the upper-right quadrant: strict-correct forced-choice decisions whose pre-answer states still show excess stress sensitivity under structured semantic stress.

\section{The \sse{} Diagnostic Lens}
\label{sec:lens}

\sse{} produces a diagnostic profile rather than a single pass/fail score. We group diagnostics into four dimensions using the fixed abbreviations \CA{}, \SO{}, \SI{}, and \BE{}: Compression--Anisotropy, Structural Organization, Selectivity--Identifiability, and Behavioral Exposure.
\begin{table*}[t]
\centering
\small
\begin{tabularx}{\textwidth}{l l X X}
\toprule
\textbf{Abbrev.} & \textbf{Dimension} & \textbf{Typical diagnostics} & \textbf{Role in this paper} \\
\midrule
\CA{} & Compression--Anisotropy & Effective rank, participation ratio, PC variance, anisotropy & Internal deformation diagnostics, not mechanism claims. \\
\SO{} & Structural Organization & Layerwise drift, peak-layer summaries, order effects & Localizes decision-state disruption. \\
\SI{} & Selectivity--Identifiability & Stress--lexical contrast, direction consistency, limited perturbation checks & Tests semantic specificity and limited perturbation sensitivity. \\
\BE{} & Behavioral Exposure & A/B accuracy, strict-correct rate, correct-only stress sensitivity, HSR & Connects internal diagnostics to behavior-only evaluation. \\
\bottomrule
\end{tabularx}
\caption{Four diagnostic dimensions in the \sse{} lens. We use the fixed abbreviations \CA{}, \SO{}, \SI{}, and \BE{} throughout the paper. The dimensions are evaluation signals, not a complete mechanistic explanation of collapse.}
\label{tab:four_dimensions}
\end{table*}
We report \CA{}, \SO{}, \SI{}, and \BE{} as a diagnostic profile rather than as a single scalar score. The main paper uses \SI{} and \BE{} for the primary evaluation claims, while \CA{} and \SO{} serve as supporting diagnostics.

\paragraph{Hidden Stress-Sensitivity Rate.}
For compact reporting, let
$C=\{\mathrm{strict\text{-}correct}(x)=1\}$ and
$U_\tau=\{\mathrm{Instab}(x)>\tau\}$. We define:
\begin{equation}
\begin{aligned}
\mathrm{HSR} &= \Pr(C \wedge U_\tau),\\
\mathrm{HSR}_{\mathrm{correct}} &= \Pr(U_\tau \mid C).
\end{aligned}
\end{equation}
Because HSR depends on threshold choice, we treat it as a summary statistic rather than the sole evidence for decoupling. Distributional and paired excess contrasts are primary.

\section{Experimental Setup}
\label{sec:setup}

\subsection{Models}
Our completed white-box runs currently include Qwen3VL-2B, Qwen3VL-4B, Qwen3VL-8B, Gemma3-4B, and InternVL3-2B \citep{qwen3vl, gemma3, internvl3}. Qwen3VL provides a within-family scale sweep; Gemma3 and InternVL3 provide complementary cross-family checks. Closed-source models are not included because behavior-only access cannot support behavior--internal decoupling analysis.

\subsection{Layer Reporting and De-Biasing}
\label{sec:layer_selection}
For each model, Table~\ref{tab:paired_excess_main} reports one diagnostic layer chosen by a fixed layer-localization rule: the peak layer from the E3 stress-specificity profile, defined as the layer with the strongest stress-to-anchor decision-state signal under stress-trial strict-correct behavior in the layerwise diagnostic pass. The selected layer is then frozen and used for both \dlex{} and \drand{}; it is not reselected to maximize \dlex{}, \drand{}, stress type, source, or intervention effect.
This reporting rule is still data-dependent. We therefore treat selected-layer values as \emph{localization diagnostics}, not as an unbiased estimate of a layer-general effect. As a limited layer-distribution check, we also report the fixed-band quantity that is available from our completed layerwise records:
\begin{equation}
\overline{S}^{\mathrm{stress}}_{i,\mathcal{L}_{\mathrm{top}}}
= \frac{1}{|\mathcal{L}_{\mathrm{top}}|}
\sum_{\ell\in\mathcal{L}_{\mathrm{top}}}
\left(1-\cos(h_{i,\ell}^{\mathrm{stress}},h_{i,\ell}^{\mathrm{pres}})\right).
\label{eq:stress_band}
\end{equation}
This is a pre-specified top-half preserve-anchored stress-displacement check. It is useful for assessing whether the selected-layer stress response is confined to one peak layer, but it does \emph{not} subtract the lexical-control displacement. We use it only as supporting evidence for distributed stress response, while the semantic-specific claim remains the selected-layer paired \dlex{} contrast in Table~\ref{tab:paired_excess_main}.

\subsection{Reporting Protocol}
For each model, condition, and diagnostic layer, we report A/B behavior, strict-correct support, correct-only stress sensitivity, paired excess contrasts, HSR where applicable, and stress-type breakdowns. We additionally report the pre-specified top-half preserve-anchored stress-displacement check in Eq.~\ref{eq:stress_band}. 
Confidence intervals for paired contrasts are computed by paired bootstrap resampling over source-item units after averaging the original and swapped orders. Low-support runs and sparse stress slices are excluded from main claims.

\section{Results}
\label{sec:results}

\subsection{Interpreting the \CA{}/\SO{}/\SI{}/\BE{} Profile}

The \CA{}/\SO{}/\SI{}/\BE{} profile is the organizing lens through which \sse{} reports multiple diagnostic views of behavior--internal decoupling. In the main results, however, these dimensions play different inferential roles. Our primary statistical claim lies at the \SI{}--\BE{} intersection: among stress-trial strict-correct items, semantic stress induces excess pre-answer decision-state displacement over lexical controls, measured by the paired \dlex{} contrast in Section~\ref{sec:primary_specificity}. \CA{} and \SO{} provide supporting context about the geometry and layerwise organization of this response; they are not combined with \SI{} and \BE{} into a single model-ranking score.
The full normalized profile values in Appendix~\ref{app:behavior_profile} are rank-normalized summaries over PC1 specificity gaps and direction-overlap terms---\emph{not} the paired \dlex{} contrast---and should be read as profile coordinates rather than directly comparable effect sizes.

\subsection{Behavioral Support, Selected-Layer Specificity, and Fixed-Band Stress Displacement}
\label{sec:primary_specificity}

Table~\ref{tab:paired_excess_main} shows original-order accuracy, swapped-order accuracy, strict-correct support, the selected diagnostic reporting layer, the lexical-control score $S_{\mathrm{lex}}$, the stress score $S_{\mathrm{stress}}$, and the paired semantic-specific excess \dlex{}. Orig./Swap/Strict are computed on the semantic-stress forced-choice trials. All internal quantities are computed only on source items that are strict-correct in both A/B orders on those stress trials.
For magnitude calibration, we report a paired standardized effect size
$d_p$ for item-level contrasts. For an item-level contrast $x_i$, such as
$x_i=\Delta_i^{\mathrm{lex}}$, we define
\begin{equation}
d_p(x)=\frac{\bar{x}}{\operatorname{sd}(x)} .
\label{eq:paired_dp}
\end{equation}
Here $\bar{x}$ is the mean of $x_i$ over strict-correct source-item units,
and $\operatorname{sd}(x)$ is the corresponding sample standard deviation.
The subscript $p$ indicates that $x_i$ is a within-item paired contrast,
computed after averaging the original and swapped caption orders. Thus
$d_p$ is a paired standardized mean difference, not an independent-groups
effect size.

\begin{table*}[t]
\centering
\scriptsize
\setlength{\tabcolsep}{2.5pt}
\begin{tabular}{lrrrrrrrrr}
\toprule
\textbf{Model} & \textbf{Orig.} & \textbf{Swap} & \textbf{Strict} & \textbf{L} & \textbf{N} & \textbf{$S_{\rm lex}$} & \textbf{$S_{\rm stress}$} & \textbf{\dlex{} [95\% CI]} & \textbf{$d_p$} \\
\midrule
Qwen3VL-2B & 0.967 & 0.861 & 0.838 & 24 & 9,661 & 0.0063 & 0.0254 & 0.0190 [0.0188,0.0192] & 2.12 \\
Qwen3VL-4B & 0.933 & 0.850 & 0.801 & 24 & 9,226 & 0.0651 & 0.1332 & 0.0681 [0.0678,0.0684] & 4.28 \\
Qwen3VL-8B & 0.904 & 0.913 & 0.839 & 32 & 9,669 & 0.1362 & 0.2655 & 0.1293 [0.1284,0.1302] & 2.85 \\
Gemma3-4B & 0.840 & 0.934 & 0.787 & 27 & 9,068 & 0.0003 & 0.0096 & 0.0093 [0.0092,0.0093] & 5.45 \\
InternVL3-2B & 0.927 & 0.903 & 0.849 & 23 & 9,785 & 0.0032 & 0.0413 & 0.0381 [0.0377,0.0384] & 2.01 \\
\bottomrule
\end{tabular}
\caption{Behavioral context and selected-layer paired specificity result. Orig./Swap are semantic-stress A/B accuracies under each option order. Strict is the stress-trial strict-correct rate: selecting the positive caption against the stress candidate in both orders. $S_{\rm lex}$ and $S_{\rm stress}$ are preserve-anchored cosine distances. \dlex{} is the paired stress-minus-lexical excess after averaging original and swapped orders; intervals are paired bootstrap 95\% CIs over source items using 3,000 resamples. $d_p$ is the paired standardized effect size defined in Eq.~\ref{eq:paired_dp}.}
\label{tab:paired_excess_main}
\end{table*}

\begin{table}[t]
\centering
\scriptsize
\setlength{\tabcolsep}{3.0pt}
\begin{tabular}{lrrr}
\toprule
\textbf{Model} & \textbf{N} & \textbf{Top-half layers} & \textbf{\sStressBand{} [95\% CI]} \\
\midrule
Qwen3VL-2B & 9,661 & 14--27 & 0.03031 [0.03013,0.03050] \\
Qwen3VL-4B & 9,226 & 18--35 & 0.06394 [0.06377,0.06411] \\
Qwen3VL-8B & 9,669 & 18--35 & 0.07665 [0.07648,0.07681] \\
Gemma3-4B & 9,068 & 17--33 & 0.00859 [0.00856,0.00862] \\
InternVL3-2B & 9,785 & 14--27 & 0.01701 [0.01687,0.01713] \\
\bottomrule
\end{tabular}
\caption{Pre-specified top-half preserve-anchored stress-displacement check. \sStressBand{} averages the stress--preserve cosine distance over the top half of decoder layers within item before bootstrapping over source items. Because this quantity is a raw distance and does not subtract the lexical-control displacement, we use it only as supporting evidence that the stress response is not confined to a single selected peak layer; the semantic-specific paired contrast remains \dlex{} in Table~\ref{tab:paired_excess_main}.}
\label{tab:fixed_band_stress_preserve}
\end{table}

Table~\ref{tab:paired_excess_main} shows the selected-layer pattern. Across completed high-support runs, \dlex{} is positive and its paired-bootstrap confidence interval excludes zero. 
Table~\ref{tab:fixed_band_stress_preserve} adds a limited top-half stress-displacement check. This fixed-band result does not subtract the lexical-control displacement, but it shows that preserve-anchored stress displacement remains visible when averaging across the top half of decoder layers instead of reporting only the localized peak. 
Selected-layer absolute stress scores range from $0.0096$ to $0.2655$, lexical-control scores range from $0.0003$ to $0.1362$, and paired excess values range from $0.0093$ to $0.1293$. We therefore interpret \dlex{} as an operational decision-state stress-sensitivity signal rather than as a complete proof of downstream fragility. The corresponding generic-mismatch contrast \drand{} is reported in Appendix~\ref{app:paired_excess}; it is positive for Qwen3VL-8B but negative for several other models, consistent with random negatives serving as a stronger raw-mismatch reference rather than the primary semantic-specific baseline.

\subsection{Scale Does Not Collapse the Evaluation Gap}

The Qwen3VL scale sweep shows that increasing model scale does not automatically collapse behavior--internal decoupling into ordinary accuracy. The semantic-specific contrast grows across the completed Qwen3VL runs $0.0190 \rightarrow 0.0681 \rightarrow 0.1293.$
This suggests that larger models may produce stronger or more identifiable stress-specific decision-state structure even when external forced-choice decisions are correct. We treat this as a within-family diagnostic trend over completed high-support runs rather than a scaling law.

\paragraph{Construct-validity boundary.}
A larger preserve-anchored displacement can reflect decision-state differentiation as well as fragility. We therefore avoid using hidden-state displacement as a reliability certificate and use the more literal description \emph{decision-state stress sensitivity}: a control-normalized signal that appears under stress-trial strict-correct behavior and survives lexical, margin, and overlap controls. The evidence does not prove that every high-score item will fail in free-form captioning or downstream QA; it shows that behavior-only correctness can hide a measurable semantic-specific internal response that is not reducible to ordinary lexical variation.

\paragraph{Margin and surface-overlap controls.}
Because answer confidence can influence decision-state geometry, we also test whether the primary effect survives explicit margin and surface-overlap controls. Appendix~\ref{app:surface_margin_controls} reports a pooled regression of the raw score on a stress-condition indicator, expected-logit margin, and positive--stress token Jaccard. The adjusted stress coefficient remains positive for every high-support model (0.0090--0.1409). A stricter margin-matched subset analysis, retaining the half of examples with the smallest absolute stress--control margin gap per model, also yields positive \dlex{} estimates (0.0089--0.1405). 

\begin{table}[t]
\centering
\scriptsize
\setlength{\tabcolsep}{3.2pt}
\begin{tabular}{lrrr}
\toprule
\textbf{Model} & \textbf{Raw \dlex{}} & \textbf{Adjusted stress} & \textbf{Margin-matched \dlex{}} \\
\midrule
Qwen3VL-2B & 0.0190 & 0.0131 & 0.0137 \\
Qwen3VL-4B & 0.0681 & 0.0597 & 0.0627 \\
Qwen3VL-8B & 0.1293 & 0.1409 & 0.1405 \\
Gemma3-4B & 0.0093 & 0.0090 & 0.0089 \\
InternVL3-2B & 0.0381 & 0.0105 & 0.0247 \\
\bottomrule
\end{tabular}
\caption{Confidence-control summary. The adjusted stress coefficient controls for expected-logit margin and token Jaccard; the margin-matched result recomputes \dlex{} on the half of items with the smallest stress--control margin gap. Full intervals are in Appendix~\ref{app:surface_margin_controls}.}
\label{tab:main_margin_controls}
\end{table}

\subsection{Generic Mismatch and Perturbation Checks}
\label{sec:generic_mismatch_checks}

Random negatives contextualize broad image--caption mismatch rather than define the semantic-specific claim. As expected, unrelated captions can induce large raw displacement, so \drand{} is reported separately in Appendix~\ref{app:paired_excess}; the primary specificity result remains the matched stress-minus-lexical contrast in Table~\ref{tab:paired_excess_main}. We also include a limited perturbation check in Appendix~\ref{app:perturbation_sanity}: in the completed runs, stress-sensitive directions produce larger answer-margin effects than sampled random-control directions. This provides a local functional sanity check for the measured directions, while the main evidence remains the strict-correct, paired \dlex{} displacement analysis.

\section{Conclusion}

We introduced \sse{}, an evaluation lens for studying behavior--internal decoupling in multimodal language models under structured semantic stress. 
Rather, semantic stress produces excess preserve-anchored decision-state displacement over lexical controls under stress-trial strict-correct behavior, while random negatives serve as generic mismatch controls and limited perturbation baselines. A pre-specified top-half preserve-anchored check further suggests that the stress response is not confined to a single localized layer. This scoped lens shows that forced-choice correctness is not a sufficient certificate of invariant internal decision geometry and motivates future work on layer-neutral estimates, mechanism, robustness tests, and mitigation.

\section*{Limitations}
\label{sec:limitations}

\paragraph{Construct validity and scope.}
\dlex{} measures a control-normalized geometric response in the pre-answer decision state. It should be interpreted as decision-state stress sensitivity, not as a direct measure of calibration, hallucination, or future behavioral failure. The signal may reflect semantic separation, modality conflict, confidence effects, or their interaction. Our conclusions are therefore scoped to structured object, attribute, relation, and compositional caption stress under the present controlled protocol.

\paragraph{Evaluation interface.}
The evidence comes from positive-anchored A/B forced-choice caption verification with original and swapped option orders. This interface gives a clean setting for behavior-conditioned hidden-state analysis, but free-form captioning, yes/no verification, broad VQA, multi-turn dialogue, and long-context visual reasoning may expose different dynamics.

\paragraph{Model and layer coverage.}
The behavior--internal analysis requires hidden-state access, so the main evidence is limited to open white-box models. In addition, the primary \dlex{} tables use selected-layer localization diagnostics. Our fixed-band check supports the presence of raw stress-related movement beyond a single selected layer, but it does not provide a fully de-biased lexical-subtracted layer-band estimate.

\paragraph{Metric interpretation.}
Different diagnostics capture different aspects of the representation. Raw displacement, \dlex{}, random-negative contrasts, HSR, profile scores, and perturbation effects should therefore be read as complementary views rather than as a single universal instability measure. Thresholded quantities such as HSR are especially sensitive to the chosen threshold and are used only as auxiliary summaries.

\paragraph{Substrate and responsible use.}
The base--preserve--stress substrate is derived from public English-centric image--caption resources, so results may reflect source distribution, caption style, cultural coverage, and object-category biases. The concurrent benchmark submission contributes the resource and black-box behavioral metrics; the present paper summarizes the construction and validation details needed for the hidden-state analysis. \dlex{} should be used as a scoped diagnostic signal, not as a safety, reliability, fairness, or deployment-readiness certificate.

\paragraph{Potential risks and responsible use.}
S$^3$E is intended as a research diagnostic for analyzing pre-answer decision-state behavior under controlled semantic stress, not as a deployment-time safety, fairness, or reliability certificate. A potential risk is over-interpreting high stress-sensitivity scores as evidence of downstream failure, or conversely treating low scores as a guarantee of robustness. Because the substrate is derived from English-centric image--caption resources, results may also reflect source-specific caption styles, visual categories, and cultural coverage. We therefore recommend using the diagnostic only for comparative research analysis, alongside behavioral, fairness, robustness, and human-centered evaluations when making deployment-relevant claims. The experiments use frozen public models and inference-only analysis; the environmental cost is limited to forward-pass evaluation rather than model training, and we report the compute budget in Appendix~\ref{app:implementation}.

\bibliography{custom}

\begin{thebibliography}{23}
\providecommand{\natexlab}[1]{#1}

\bibitem[{Anonymous(2026)}]{anonymous2026semanticstress}
Anonymous. 2026.
\newblock A structured semantic-stress benchmark for multimodal model evaluation.
\newblock Concurrent anonymous benchmark submission. An anonymized manuscript should be included in supplementary material.

\bibitem[{Bai et~al.(2025)Bai, Cai, Chen, Chen, Chen, Cheng, Deng, Ding, Gao, Ge et~al.}]{qwen3vl}
Shuai Bai, Yuxuan Cai, Ruizhe Chen, Keqin Chen, Xionghui Chen, Zesen Cheng, Lianghao Deng, Wei Ding, Chang Gao, Chunjiang Ge, and 1 others. 2025.
\newblock \href {https://arxiv.org/abs/2511.21631} {Qwen3-vl technical report}.
\newblock \emph{arXiv preprint arXiv:2511.21631}.

\bibitem[{Ethayarajh(2019)}]{ethayarajh2019contextual}
Kawin Ethayarajh. 2019.
\newblock \href {https://doi.org/10.18653/v1/D19-1006} {How contextual are contextualized word representations? comparing the geometry of {BERT}, {ELMo}, and {GPT}-2 embeddings}.
\newblock In \emph{Proceedings of the 2019 Conference on Empirical Methods in Natural Language Processing and the 9th International Joint Conference on Natural Language Processing (EMNLP-IJCNLP)}, pages 55--65, Hong Kong, China. Association for Computational Linguistics.

\bibitem[{Fu et~al.(2023)Fu, Chen, Shen, Qin, Zhang, Lin, Yang, Zheng, Li, Sun, Wu, and Ji}]{fu2023mme}
Chaoyou Fu, Peixian Chen, Yunhang Shen, Yulei Qin, Mengdan Zhang, Xu~Lin, Jinrui Yang, Xiawu Zheng, Ke~Li, Xing Sun, Yunsheng Wu, and Rongrong Ji. 2023.
\newblock Mme: A comprehensive evaluation benchmark for multimodal large language models.
\newblock \emph{arXiv preprint arXiv:2306.13394}.

\bibitem[{Geiger et~al.(2021)Geiger, Lu, Icard, and Potts}]{geiger2021causal}
Atticus Geiger, Hanson Lu, Thomas Icard, and Christopher Potts. 2021.
\newblock Causal abstractions of neural networks.
\newblock In \emph{Advances in Neural Information Processing Systems}.

\bibitem[{{Gemma Team}(2025)}]{gemma3}
{Gemma Team}. 2025.
\newblock \href {https://arxiv.org/abs/2503.19786} {Gemma 3 technical report}.
\newblock \emph{arXiv preprint arXiv:2503.19786}.

\bibitem[{Guan et~al.(2024)Guan, Liu, Wu, Xian, Li, Liu, Wang, Chen, Huang, Yacoob, Manocha, and Zhou}]{guan2024hallusionbench}
Tianrui Guan, Fuxiao Liu, Xiyang Wu, Ruiqi Xian, Zongxia Li, Xiaoyu Liu, Xijun Wang, Lichang Chen, Furong Huang, Yaser Yacoob, Dinesh Manocha, and Tianyi Zhou. 2024.
\newblock \href {https://openaccess.thecvf.com/content/CVPR2024/html/Guan_HallusionBench_An_Advanced_Diagnostic_Suite_for_Entangled_Language_Hallucination_and_CVPR_2024_paper.html} {Hallusionbench: An advanced diagnostic suite for entangled language hallucination and visual illusion in large vision-language models}.
\newblock In \emph{Proceedings of the IEEE/CVF Conference on Computer Vision and Pattern Recognition (CVPR)}.

\bibitem[{Hewitt and Manning(2019)}]{hewitt2019structural}
John Hewitt and Christopher~D. Manning. 2019.
\newblock A structural probe for finding syntax in word representations.
\newblock In \emph{Proceedings of the 2019 Conference of the North American Chapter of the Association for Computational Linguistics: Human Language Technologies}.

\bibitem[{Hsieh et~al.(2023)Hsieh, Zhang, Ma, Kembhavi, and Krishna}]{hsieh2023sugarcrepe}
Cheng-Yu Hsieh, Jieyu Zhang, Zixian Ma, Aniruddha Kembhavi, and Ranjay Krishna. 2023.
\newblock \href {https://proceedings.neurips.cc/paper_files/paper/2023/hash/63461de0b4cb760fc498e85b18a7fe81-Abstract-Datasets_and_Benchmarks.html} {Sugarcrepe: Fixing hackable benchmarks for vision-language compositionality}.
\newblock In \emph{Advances in Neural Information Processing Systems}, volume~36.

\bibitem[{Kornblith et~al.(2019)Kornblith, Norouzi, Lee, and Hinton}]{kornblith2019similarity}
Simon Kornblith, Mohammad Norouzi, Honglak Lee, and Geoffrey Hinton. 2019.
\newblock Similarity of neural network representations revisited.
\newblock In \emph{Proceedings of the 36th International Conference on Machine Learning}.

\bibitem[{Li et~al.(2023)Li, Du, Zhou, Wang, Zhao, and Wen}]{li2023pope}
Yifan Li, Yifan Du, Kun Zhou, Jinpeng Wang, Xin Zhao, and Ji-Rong Wen. 2023.
\newblock \href {https://doi.org/10.18653/v1/2023.emnlp-main.20} {Evaluating object hallucination in large vision-language models}.
\newblock In \emph{Proceedings of the 2023 Conference on Empirical Methods in Natural Language Processing}, pages 292--305, Singapore. Association for Computational Linguistics.

\bibitem[{Liang et~al.(2025)Liang, Chen, Jiao, Liang, Liu, Zhang, Hu, and Cao}]{liang2025intramodal}
Jiawei Liang, Ruoyu Chen, Xianghao Jiao, Siyuan Liang, Shiming Liu, Qunli Zhang, Zheng Hu, and Xiaochun Cao. 2025.
\newblock \href {https://arxiv.org/abs/2509.22415} {Explaining multimodal {LLM}s via intra-modal token interactions}.
\newblock \emph{Preprint}, arXiv:2509.22415.

\bibitem[{Liu et~al.(2024)Liu, Duan, Zhang, Li, Zhang, Zhao, Yuan, Wang, He, Liu, Chen, and Lin}]{liu2024mmbench}
Yuanzhan Liu, Haodong Duan, Yuanhan Zhang, Bo~Li, Songyang Zhang, Wangbo Zhao, Yike Yuan, Jiaqi Wang, Conghui He, Ziwei Liu, Kai Chen, and Dahua Lin. 2024.
\newblock Mmbench: Is your multi-modal model an all-around player?
\newblock In \emph{European Conference on Computer Vision}.

\bibitem[{Meng et~al.(2022)Meng, Bau, Andonian, and Belinkov}]{meng2022locating}
Kevin Meng, David Bau, Alex Andonian, and Yonatan Belinkov. 2022.
\newblock Locating and editing factual associations in gpt.
\newblock In \emph{Advances in Neural Information Processing Systems}.

\bibitem[{Raghu et~al.(2017)Raghu, Gilmer, Yosinski, and Sohl-Dickstein}]{raghu2017svcca}
Maithra Raghu, Justin Gilmer, Jason Yosinski, and Jascha Sohl-Dickstein. 2017.
\newblock Svcca: Singular vector canonical correlation analysis for deep learning dynamics and interpretability.
\newblock In \emph{Advances in Neural Information Processing Systems}.

\bibitem[{Rimsky et~al.(2024)Rimsky, Gabrieli, Schulz, Tong, Hubinger, and Turner}]{rimsky2024steering}
Nina Rimsky, Nathan Gabrieli, Julian Schulz, Meg Tong, Evan Hubinger, and Alexander~Matt Turner. 2024.
\newblock Steering llama 2 via contrastive activation addition.
\newblock In \emph{Proceedings of the 62nd Annual Meeting of the Association for Computational Linguistics}.

\bibitem[{Rogers et~al.(2020)Rogers, Kovaleva, and Rumshisky}]{rogers2020primer}
Anna Rogers, Olga Kovaleva, and Anna Rumshisky. 2020.
\newblock A primer in bertology: What we know about how bert works.
\newblock \emph{Transactions of the Association for Computational Linguistics}, 8:842--866.

\bibitem[{Rohrbach et~al.(2018)Rohrbach, Hendricks, Burns, Darrell, and Saenko}]{rohrbach2018object}
Anna Rohrbach, Lisa~Anne Hendricks, Kaylee Burns, Trevor Darrell, and Kate Saenko. 2018.
\newblock \href {https://doi.org/10.18653/v1/D18-1437} {Object hallucination in image captioning}.
\newblock In \emph{Proceedings of the 2018 Conference on Empirical Methods in Natural Language Processing}, pages 4035--4045, Brussels, Belgium. Association for Computational Linguistics.

\bibitem[{Thrush et~al.(2022)Thrush, Jiang, Bartolo, Singh, Williams, Kiela, and Ross}]{thrush2022winoground}
Tristan Thrush, Ryan Jiang, Max Bartolo, Amanpreet Singh, Adina Williams, Douwe Kiela, and Candace Ross. 2022.
\newblock \href {https://doi.org/10.1109/CVPR52688.2022.00517} {Winoground: Probing vision and language models for visio-linguistic compositionality}.
\newblock In \emph{Proceedings of the IEEE/CVF Conference on Computer Vision and Pattern Recognition (CVPR)}, pages 5238--5248.

\bibitem[{Vig et~al.(2020)Vig, Gehrmann, Belinkov, Qian, Nevo, Singer, and Shieber}]{vig2020causal}
Jesse Vig, Sebastian Gehrmann, Yonatan Belinkov, Sharon Qian, Daniel Nevo, Yaron Singer, and Stuart Shieber. 2020.
\newblock Investigating gender bias in language models using causal mediation analysis.
\newblock In \emph{Advances in Neural Information Processing Systems}.

\bibitem[{Yu et~al.(2024)Yu, Yang, Li, Wang, Lin, Liu, Wang, and Wang}]{yu2024mmvet}
Weihao Yu, Zhengyuan Yang, Linjie Li, Jianfeng Wang, Kevin Lin, Zicheng Liu, Xinchao Wang, and Lijuan Wang. 2024.
\newblock Mm-vet: Evaluating large multimodal models for integrated capabilities.
\newblock In \emph{Proceedings of the 41st International Conference on Machine Learning}.

\bibitem[{Y{\"u}ksekg{\"o}n{\"u}l et~al.(2023)Y{\"u}ksekg{\"o}n{\"u}l, Bianchi, Kalluri, Jurafsky, and Zou}]{yuksekgonul2023aro}
Mert Y{\"u}ksekg{\"o}n{\"u}l, Federico Bianchi, Pratyusha Kalluri, Dan Jurafsky, and James Zou. 2023.
\newblock \href {https://openreview.net/forum?id=KRLUvxh8uaX} {When and why vision-language models behave like bags-of-words, and what to do about it?}
\newblock In \emph{The Eleventh International Conference on Learning Representations (ICLR)}.

\bibitem[{Zhu et~al.(2025)Zhu, Wang, Chen, Liu, Ye, Gu, Duan, Tian, Su, Shao, Gao, Cui et~al.}]{internvl3}
Jinguo Zhu, Weiyun Wang, Zhe Chen, Zhaoyang Liu, Shenglong Ye, Lixin Gu, Yuchen Duan, Hao Tian, Weijie Su, Jie Shao, Zhangwei Gao, Erfei Cui, and 1 others. 2025.
\newblock \href {https://arxiv.org/abs/2504.10479} {Internvl3: Exploring advanced training and test-time recipes for open-source multimodal models}.
\newblock \emph{arXiv preprint arXiv:2504.10479}.

\end{thebibliography}

\appendix

\section{\sse{} Protocol and Prompt Templates}
\label{app:protocol}

\sse{} uses a positive-anchored A/B forced-choice interface rather than free-form caption generation or broad VQA. Each evaluation unit contains an image $I$, an image-supported positive caption $c^+$, and a candidate caption $c$ drawn from one of the semantic-stress or control conditions. The prompt asks the model to choose which caption better matches the image.

\paragraph{Prompt template.}
The shared prompt template is:
\begin{quote}
\small
\texttt{<image>}\\
\texttt{Which caption better matches the image? Answer only A or B.}\\
\texttt{A: ``\{caption\_A\}''}\\
\texttt{B: ``\{caption\_B\}''}
\end{quote}
No free-form caption is generated. We use only the forward-pass logits for the A/B decision and hidden states at the pre-answer decision site.

\paragraph{Original and swapped orders.}
For each candidate condition, we evaluate both option orders:
\begin{align}
\text{original:} \quad & A=c^+, \quad B=c,\\
\text{swapped:} \quad & A=c, \quad B=c^+.
\end{align}
For negative candidates, the expected answer is $A$ in the original order and $B$ in the swapped order. We define the main \emph{strict-correct} filter on semantic-stress trials: a source item passes when the model selects the positive caption against the stress candidate in both option orders. This filter controls simple A/B position bias before internal-state diagnostics are computed. Preserve and lexical controls are meaning-preserving alternatives and are not assigned forced behavioral correctness labels; their hidden states are used as matched internal controls for source items that pass the stress-trial strict-correct filter.

\paragraph{Condition families.}
The protocol distinguishes five roles: the positive reference, preserve controls, semantic stress, lexical controls, and random negatives. Semantic-stress candidates change a targeted semantic factor; the positive caption provides the reference state for Eq.~\ref{eq:primary_score}; preserve and lexical controls are distinct meaning-preserving comparison conditions; random negatives are generic caption-mismatch controls.

\section{Anonymous Benchmark Substrate}
\label{app:benchmark_substrate}

This appendix gives the substrate details needed to reproduce and interpret the \sse{} analysis while preserving anonymous-review separation from the concurrent benchmark submission.

\paragraph{Relation to concurrent work.}
The structured semantic-stress candidate set is drawn from a concurrent anonymous benchmark submission~\citep{anonymous2026semanticstress}. We cite that manuscript anonymously and separate its scope from the present paper. The benchmark submission is responsible for the public-source-derived base--preserve--stress resource, black-box behavioral metrics, and resource validation. The present paper is responsible for the \sse{} evaluation lens, pre-answer decision-state scoring, stress-trial strict-correct behavior matching, and control-normalized hidden-state contrasts.

\paragraph{Public source pools.}
The source items are derived from Winoground, SugarCrepe, and ARO. The raw candidate snapshot contains 426,278 stress variants; the feasible three-source pool contains 418,969 variants over 30,531 images. The curated benchmark core contains 11,522 stress variants over 7,838 images and is the data source for the hidden-state experiments reported here. The benchmark core reduces source dominance relative to the raw candidate snapshot: ARO contributes 55.0\%, SugarCrepe 39.0\%, and Winoground 6.0\% after curation. The concurrent benchmark also defines a smaller 4,000-variant matched analysis subset over 3,401 images for black-box behavioral probes. These counts are reported so that the current paper is not dependent on a vague external substrate description.

\paragraph{Canonical item format.}
The benchmark-level item has a base--preserve--stress structure: a source base caption, a meaning-preserving preserve caption, and a meaning-changing stress caption. In the present paper, the image-supported positive/preserve caption supplies $c^+$ and the preserve anchor notation $c^{\mathrm{pres}}$ used in the scoring equations. The lexical control $c^{\mathrm{lex}}$ is a separate meaning-preserving surface-control caption, not the preserve anchor. The stress caption introduces a targeted unsupported perturbation. Stress variants are labeled as object, attribute, relation, or compositional; compositional cases may be split into order-compositional and high-confusion sub-slices for reporting. Operation metadata include entity substitution, attribute injection, relation modification, attachment shift, lexical fallback, and compositional reordering.

\paragraph{Conversion into \sse{} roles.}
The \sse{} record used for hidden-state analysis contains
\begin{equation}
(I, c^+, c^{\mathrm{pres}}, c^{\mathrm{lex}}, c^{\mathrm{stress}}, c^{\mathrm{rand}}, t),
\end{equation}
where $I$ is the image, $c^+$ is the image-supported positive caption, $c^{\mathrm{pres}}$ is a meaning-preserving preserve control, $c^{\mathrm{lex}}$ is a distinct meaning-preserving surface or lexical control, $c^{\mathrm{stress}}$ is a meaning-changing semantic foil, $c^{\mathrm{rand}}$ is a random negative sampled from another item, and $t$ is the stress-type label. The positive, preserve, and stress roles are inherited from the base--preserve--stress substrate; lexical and random conditions are control roles used for the internal analysis.

\paragraph{Filtering and matching principles.}
Items are retained only when the source example can be mapped into the required positive, preserve, lexical, stress, and random roles. The conversion excludes degenerate cases such as identical candidate strings, missing image or caption fields, and random negatives drawn from the same source item. Curation applies source/readiness filtering, rule-based severity scoring, duplicate reduction, per-image caps, source balancing, and slice-aware sampling. The benchmark manuscript reports a human pilot on 120 stratified examples with five annotators per item; majority pass rates are high for preserve validity and stress validity, while single-factor and severity judgments are lower and therefore treated as coarser audit signals rather than perfect labels.

\paragraph{Behavioral-instability labels.}
The concurrent benchmark also defines black-box behavioral probes on its 4,000-variant matched analysis subset: order sensitivity, prompt sensitivity, repeat sensitivity, and any-instability. These are behavioral labels computed from mapped preserve/stress decisions under interface perturbations. They do not claim access to internal states. The present paper does not use these labels or that smaller subset in the reported hidden-state analysis; they are described only to delineate the benchmark substrate from the white-box decision-state analysis.

\paragraph{Anonymization and reproducibility.}
Following anonymous-review practice for related concurrent submissions, the final supplementary package should include an anonymized copy of the benchmark manuscript and an anonymized evaluation subset or subset description with public source identifiers where license-compatible. Supplementary files should exclude author names, institutional paths, non-anonymous repositories, and local filesystem metadata. The main paper remains self-contained with respect to the experiments reported here: it identifies the public sources, defines all candidate roles, prompt format, correctness filters, diagnostic dimensions, and statistical contrasts needed to interpret the results.

\paragraph{Artifact licenses and access.}
Our substrate is derived from public research artifacts rather than newly collected private data. SugarCrepe code/data are distributed through the official repository, which lists an MIT license for the repository; the benchmark uses COCO images, so users should also comply with COCO's terms. Winoground is accessed through Hugging Face under its dataset access conditions, including research-use terms, and we do not redistribute Winoground images. ARO-derived items are accessed from the public Hugging Face dataset releases; where an upstream license is not explicitly specified, we treat the data as research-use artifacts and redistribute only derived metadata/source identifiers when license-compatible. Any released S$^3$E code will be distributed under MIT license, while derived annotations or evaluation metadata will be released under CC BY 4.0 license only to the extent compatible with upstream artifact terms. The release package will include source identifiers, scripts, and documentation sufficient to reconstruct the evaluation where redistribution of original images or captions is restricted.

\paragraph{Artifact use and intended use.}
The existing artifacts used in this paper are public research benchmarks for evaluating image--text grounding, compositionality, and multimodal model behavior. We use Winoground, SugarCrepe, and ARO-derived examples only for research evaluation and diagnostic analysis, which is consistent with their benchmark-oriented intended use. We do not use these artifacts to train or fine-tune models, to build a deployed system, or to make claims about individual people in images. For any derived artifacts released with this paper, the intended use is research-only evaluation of multimodal models under controlled semantic stress. Redistribution will be limited to source identifiers, derived annotations, prompts, scripts, and metadata where compatible with upstream access conditions; original images or captions will not be redistributed when upstream licenses or access terms do not permit redistribution.

\paragraph{Personally identifying and offensive content.}
We do not collect new images, personal information, demographic attributes, names, contact information, or user-generated private text. The data are inherited from existing public image--caption research benchmarks, and our added annotations describe semantic-stress roles such as object, attribute, relation, and compositional perturbations rather than identities of people. Because some upstream images may contain people, we avoid adding identity labels, face annotations, or person-specific demographic inferences. During curation, we remove degenerate records such as missing images or captions, identical candidate strings, and invalid role mappings; the human audit checks preserve validity, stress validity, naturalness, and single-factor control rather than attempting to identify people. The benchmark is not designed to contain offensive content, but we do not claim an exhaustive audit of all upstream image pixels or captions for offensive material. If offensive or personally identifying content is found in the released metadata, it should be reported and removed from future versions.

\paragraph{Artifact documentation.}
The artifact documentation specifies the source benchmarks, candidate roles, stress types, operation metadata, severity labels, source family, diagnostic slices, and evaluation protocol used in this paper. The released benchmark core used here contains 11,522 stress variants over 7,838 images, derived from Winoground, SugarCrepe, and ARO. The documented stress categories include object, attribute, relation, and compositional/order perturbations, and the evaluation records define positive, preserve, lexical, stress, and random roles for the S$^3$E diagnostic. The documentation also reports curation steps, source balancing, duplicate reduction, per-image caps, validity filtering, and human-audit summary statistics. The artifact is English-centric and image--caption based, so its coverage is limited by the source distributions, visual categories, caption styles, and cultural assumptions of the upstream benchmarks.

\section{Metric Definitions for \CA{}, \SO{}, \SI{}, and \BE{}}
\label{app:metric_definitions}

The four-dimensional profile is intended as an evaluation summary rather than a mechanistic decomposition. All diagnostics are computed on pre-answer decision states unless otherwise specified.

\paragraph{\CA{}: Compression--Anisotropy.}
\CA{} summarizes whether stress-sensitive decision states show concentration or compression. Typical ingredients include effective-rank change, participation ratio, principal-component variance concentration, and anisotropy. For a set of hidden states with covariance eigenvalues $\lambda_j$, a standard effective-rank form is
\begin{equation}
\mathrm{erank}=\exp\left(-\sum_j p_j \log p_j\right), \quad p_j=\frac{\lambda_j}{\sum_k \lambda_k}.
\end{equation}
Lower effective rank or stronger top-PC concentration indicates a more compressed or anisotropic profile. We treat these as diagnostic signals, not as a standalone causal mechanism.

\paragraph{\SO{}: Structural Organization.}
\SO{} summarizes whether the diagnostic signal is organized across layers and option orders. It uses layerwise profiles, selected peak layers, original--swapped aggregation, and split-half consistency. A high \SO{} profile indicates that the signal is not an isolated one-off perturbation at an arbitrary layer.

\paragraph{\SI{}: Selectivity--Identifiability.}
\SI{} is computed from condition-level PC1 specificity gaps (stress vs. lexical and stress vs. preserve) and a direction-overlap separability term (split-half stress overlap minus stress–lexical direction overlap). These quantities characterize whether stress-induced representational variance is concentrated and separable from control directions at the profile level. \SI{} is distinct from the paired \dlex{} contrast in Eq. 4: the latter is an item-level paired distance, while \SI{} aggregates condition-level geometric summaries. Direction consistency and limited perturbation checks, where available, are also treated as auxiliary \SI{} evidence.

\paragraph{\BE{}: Behavioral Exposure.}
\BE{} connects internal diagnostics to behavior-only evaluation. It includes pair accuracy, strict-correct rate, correct-only stress sensitivity, and hidden stress-sensitivity summaries such as HSR. \BE{} is not a replacement for internal diagnostics; it identifies whether an internal signal is hidden behind correct behavior or exposed as behavioral failure.

\section{Control-Normalized Contrasts}
\label{app:contrasts}

We do not define stress sensitivity as arbitrary hidden-state variation across different captions. Let $S(h)$ be an internal diagnostic score computed from the pre-answer decision state relative to the positive-caption reference. The preserve specificity contrast is
\begin{equation}
\Delta^{\mathrm{pres}}_{i,o}=S(h_{i,o,\mathrm{stress}})-S(h_{i,o,\mathrm{pres}}),
\end{equation}
and the lexical specificity contrast is
\begin{equation}
\Delta^{\mathrm{lex}}_{i,o}=S(h_{i,o,\mathrm{stress}})-S(h_{i,o,\mathrm{lex}}),
\end{equation}
computed only on source items that pass the semantic-stress strict-correct filter. \dpres{} tests whether meaning-changing stress produces excess diagnostic signal beyond a meaning-preserving reformulation, while \dlex{} tests whether the signal exceeds ordinary lexical or caption-surface variation.

The generic mismatch contrast is
\begin{equation}
\Delta^{\mathrm{rand}}_{i,o}=S(h_{i,o,\mathrm{stress}})-S(h_{i,o,\mathrm{random}}).
\end{equation}
Random negatives are a stronger mismatch reference, not the primary semantic-specific baseline. Therefore, \drand{} is reported as a control diagnostic and is not expected to be positive for every raw metric or every model.

\section{Perturbation Sanity Check}
\label{app:perturbation_sanity}

We use the completed E5 perturbation runs to test whether directions selected from stress-sensitive decision-state responses have a larger local answer-margin footprint than sampled random-control directions. Table~\ref{tab:perturbation_sanity} reports the excess margin perturbation of stress-sensitive directions over random-control directions in the completed runs. Because coverage is limited, this check is used as a supplementary functional sanity check rather than as part of the primary paired-displacement claim.

\begin{table}[t]
\centering
\scriptsize
\setlength{\tabcolsep}{3.2pt}
\begin{tabular}{@{}lrrrr@{}}
\toprule
\textbf{Run} & \textbf{N} & \textbf{$\Delta_{\mathrm{E5}}$} & \textbf{95\% CI} & \textbf{$P(\Delta>0)$} \\
\midrule
Qwen-4B L27 & 400 & 9.34 & [8.68,10.03] & 99.2 \\
Qwen-4B L28 & 400 & 8.61 & [8.07,9.17] & 100.0 \\
Gemma-4B L27 & 400 & 15.90 & [14.61,17.22] & 100.0 \\
\bottomrule
\end{tabular}
\caption{Supplementary E5 perturbation sanity check. \dE{} is the excess answer-margin perturbation of stress-sensitive directions over sampled random-control directions in the completed runs.}
\label{tab:perturbation_sanity}
\end{table}

\section{Surface-Text and Margin Controls}
\label{app:surface_margin_controls}

These controls quantify surface caption overlap and answer-margin associations. They are not used as independent causal evidence; they document possible nuisance factors and make explicit that the paper does not define stress sensitivity as arbitrary caption-induced variation. Because the preserve anchor is the positive reference caption rather than an independent surface-control candidate, Table~\ref{tab:app_surface_similarity} reports stress-candidate overlap with the positive caption. Tables~\ref{tab:app_margin_surface_regression} and~\ref{tab:app_margin_matched} then test whether the primary contrast remains positive after controlling for expected-logit margin and overlap.

\begin{table*}[!htb]
\centering
\small
\begin{tabular}{lrrrr}
\toprule
\textbf{Slice} & \textbf{N} & \textbf{Token Jaccard} & \textbf{Token F1} & \textbf{Length ratio} \\
\midrule
Overall & 11522 & 0.630 $\pm$ 0.090 & 0.769 $\pm$ 0.071 & 0.812 $\pm$ 0.107 \\
Attribute & 3832 & 0.641 $\pm$ 0.100 & 0.776 $\pm$ 0.079 & 0.779 $\pm$ 0.074 \\
Compositional & 96 & 0.553 $\pm$ 0.234 & 0.681 $\pm$ 0.211 & 0.578 $\pm$ 0.204 \\
Object & 3573 & 0.627 $\pm$ 0.071 & 0.768 $\pm$ 0.053 & 0.761 $\pm$ 0.045 \\
Relation & 4021 & 0.625 $\pm$ 0.088 & 0.766 $\pm$ 0.068 & 0.894 $\pm$ 0.115 \\
\bottomrule
\end{tabular}
\caption{Surface similarity between the positive caption and stress candidate. Values are means $\pm$ standard deviations.}
\label{tab:app_surface_similarity}
\end{table*}

\begin{table*}[!htb]
\centering
\scriptsize
\setlength{\tabcolsep}{3pt}
\begin{tabular}{lrrrrrrr}
\toprule
\textbf{Model} & \textbf{N} & \textbf{Raw \dlex{}} & \textbf{Adj. stress} & \textbf{95\% CI} & \textbf{$\beta_{\mathrm{margin}}$} & \textbf{$\beta_{\mathrm{jac}}$} & \textbf{$R^2$} \\
\midrule
Qwen3VL-2B & 9,661 & 0.0190 & 0.0131 & [0.0129,0.0132] & 0.0025 & 0.0012 & 0.884 \\
Qwen3VL-4B & 9,226 & 0.0681 & 0.0597 & [0.0591,0.0603] & 0.0010 & -0.0257 & 0.855 \\
Qwen3VL-8B & 9,669 & 0.1293 & 0.1409 & [0.1398,0.1420] & -0.0038 & -0.1298 & 0.663 \\
Gemma3-4B & 9,068 & 0.0093 & 0.0090 & [0.0090,0.0091] & 0.0000 & -0.0033 & 0.939 \\
InternVL3-2B & 9,785 & 0.0381 & 0.0105 & [0.0101,0.0108] & 0.0054 & 0.0010 & 0.878 \\
\bottomrule
\end{tabular}
\caption{Margin and surface-overlap adjusted analysis. We fit a pooled model to stress and lexical scores, $S=\alpha+\beta_{s}\mathbf{1}[\mathrm{stress}]+\beta_m\mathrm{margin}+\beta_j\mathrm{Jaccard}+\epsilon$, with paired bootstrap intervals over source items. The adjusted stress coefficient remains positive across completed high-support models.}
\label{tab:app_margin_surface_regression}
\end{table*}

\begin{table}[h]
\centering
\scriptsize
\setlength{\tabcolsep}{3pt}
\begin{tabular}{lrrrrr}
\toprule
\textbf{Model} & \textbf{N} & \textbf{$|\Delta_m|$ thr.} & \textbf{\dlex{}} & \textbf{95\% CI} & \textbf{$P(\Delta>0)$} \\
\midrule
Qwen3VL-2B & 4,849 & 2.344 & 0.0137 & [0.0136,0.0139] & 100.0 \\
Qwen3VL-4B & 4,627 & 8.750 & 0.0627 & [0.0622,0.0632] & 100.0 \\
Qwen3VL-8B & 4,841 & 3.719 & 0.1405 & [0.1393,0.1417] & 100.0 \\
Gemma3-4B & 4,579 & 6.562 & 0.0089 & [0.0088,0.0089] & 100.0 \\
InternVL3-2B & 4,926 & 4.969 & 0.0247 & [0.0244,0.0251] & 100.0 \\
\bottomrule
\end{tabular}
\caption{Margin-matched robustness check. For each model, we retain the half of source items with the smallest absolute stress--control expected-logit margin gap $|\Delta_m|$ and recompute \dlex{}. The semantic-specific contrast remains positive.}
\label{tab:app_margin_matched}
\end{table}

\section{Strict-Correct Behavior and Diagnostic Profiles}
\label{app:behavior_profile}

Table~\ref{tab:app_behavior_profile} reports the completed model-level behavior summary and normalized four-dimensional profiles. The strict-correct rate column is computed on semantic-stress trials and is the behavioral support used for correct-only internal analyses.

\begin{table*}[h]
\centering
\small
\begin{tabular}{lrrrrrr}
\toprule
\textbf{Model} & \textbf{Pair Acc.} & \textbf{Strict-correct} & \textbf{\CA{}} & \textbf{\SO{}} & \textbf{\SI{}} & \textbf{\BE{}} \\
\midrule
Qwen3VL-2B & 0.914 & 0.838 & 0.000 & 0.005 & 0.499 & 0.173 \\
Qwen3VL-4B & 0.891 & 0.801 & 0.921 & 0.148 & 0.767 & 0.780 \\
Qwen3VL-8B & 0.908 & 0.839 & 0.478 & 0.370 & 0.000 & 0.162 \\
Gemma3-4B & 0.887 & 0.787 & 0.092 & 0.000 & 0.332 & 1.000 \\
InternVL3-2B & 0.915 & 0.849 & 1.000 & 1.000 & 1.000 & 0.000 \\
\bottomrule
\end{tabular}
\caption{Completed semantic-stress behavior summary and normalized \CA{}/\SO{}/\SI{}/\BE{} diagnostic profiles. These profile values are rank-normalized summaries over heterogeneous submetrics and are not the primary statistical claim. They can therefore differ from the primary paired \dlex{} results in Table~\ref{tab:paired_excess_main}.}
\label{tab:app_behavior_profile}
\end{table*}

\section{Position-Balance Diagnostics}
\label{app:position_balance}

The original and swapped orders expose A/B position effects before strict-correct filtering. For the main analysis, strict correctness requires the model to select the positive caption against the semantic-stress candidate in both orders, which removes trials explainable by a single-order position preference.

\begin{table*}[h]
\centering
\small
\begin{tabular}{lrrrrr}
\toprule
\textbf{Model} & \textbf{Orig. acc.} & \textbf{Swap acc.} & \textbf{Orig. A-rate} & \textbf{Swap A-rate} & \textbf{Gap} \\
\midrule
Qwen3VL-2B & 0.967 & 0.861 & 0.967 & 0.139 & 0.106 \\
Qwen3VL-4B & 0.933 & 0.850 & 0.933 & 0.150 & 0.083 \\
Qwen3VL-8B & 0.904 & 0.913 & 0.904 & 0.087 & -0.009 \\
Gemma3-4B & 0.840 & 0.934 & 0.840 & 0.066 & -0.094 \\
InternVL3-2B & 0.927 & 0.903 & 0.927 & 0.097 & 0.024 \\
\bottomrule
\end{tabular}
\caption{Original/swapped semantic-stress behavior before strict-correct filtering. The final diagnostic analysis uses stress-trial strict-correct source items rather than either order alone.}
\label{tab:app_position_bias}
\end{table*}

\section{Full Paired Excess Results}
\label{app:paired_excess}

The main paper reports compact paired excess results with paired bootstrap intervals for the primary semantic-specific contrast. This appendix expands those results by source and stress type. We include completed high-support models only; incomplete or low-support preliminary rows are omitted.

\subsection{Overall Paired Excess}

Table~\ref{tab:app_paired_overall} reports the aggregate paired-excess diagnostic for each completed high-support model. Across all models, \dlex{} is positive, indicating that semantic-conflict candidates induce larger preserve-anchored decision-state displacement than lexical controls under matched correct behavior. In contrast, \drand{} is not consistently positive, confirming that random mismatches serve as a generic mismatch reference rather than the primary semantic-specific baseline.

\begin{table*}[!htb]
\centering
\scriptsize
\setlength{\tabcolsep}{3pt}
\begin{tabular}{lrrrrrrrr}
\toprule
\textbf{Model} & \textbf{L} & \textbf{N} & \textbf{\dlex{} mean} & \textbf{\dlex{} CI} & \textbf{$\%\dlex{}>0$} & \textbf{\drand{} mean} & \textbf{\drand{} CI} & \textbf{$\%\drand{}>0$} \\
\midrule
Gemma3-4B & 27 & 9,068 & 0.0093 & [0.0092,0.0093] & 100.0 & -0.0016 & [-0.0016,-0.0015] & 12.8 \\
InternVL3-2B & 23 & 9,785 & 0.0381 & [0.0377,0.0384] & 100.0 & -0.0399 & [-0.0403,-0.0394] & 3.2 \\
Qwen3VL-2B & 24 & 9,661 & 0.0190 & [0.0188,0.0192] & 100.0 & -0.0156 & [-0.0158,-0.0154] & 5.3 \\
Qwen3VL-4B & 24 & 9,226 & 0.0681 & [0.0678,0.0684] & 100.0 & -0.0147 & [-0.0149,-0.0145] & 8.6 \\
Qwen3VL-8B & 32 & 9,669 & 0.1293 & [0.1284,0.1302] & 99.9 & 0.0295 & [0.0288,0.0301] & 85.3 \\
\bottomrule
\end{tabular}
\caption{Overall paired excess results for completed high-support models. \dlex{} is the primary semantic-specific contrast; \drand{} is the generic mismatch contrast. Intervals are paired bootstrap 95\% confidence intervals over source-item units after original/swapped order averaging.}
\label{tab:app_paired_overall}
\end{table*}

\subsection{Source/Slice-Level Breakdown}
Table~\ref{tab:app_paired_source} breaks the paired-excess diagnostic down by source pool and analysis slice. The goal is not to rank source datasets, but to check whether the aggregate \dlex{} signal is driven by a single source. The positive \dlex{} pattern appears across Winoground, SugarCrepe, and ARO-derived slices, suggesting that the semantic-specific displacement effect is not confined to one benchmark substrate.

\begin{table*}[h]
\centering
\tiny
\begin{tabular}{llrrrrrr}
\toprule
\textbf{Model} & \textbf{Source/Slice} & \textbf{Layer} & \textbf{Strict N} & \textbf{\dlex{} mean} & \textbf{\drand{} mean} & \textbf{\drand{} 95\% CI} & \textbf{$\%>0$} \\
\midrule
Qwen3VL-2B & aro & 24 & 3,611 & 0.0171 & -0.0144 & [-0.0147, -0.0141] & 4.3 \\
Qwen3VL-2B & ARO-relation slice & 24 & 1,701 & 0.0184 & -0.0118 & [-0.0123, -0.0113] & 13.6 \\
Qwen3VL-2B & sugarcrepe & 24 & 3,763 & 0.0214 & -0.0194 & [-0.0197, -0.0191] & 1.1 \\
Qwen3VL-2B & winoground & 24 & 586 & 0.0170 & -0.0097 & [-0.0104, -0.0090] & 14.7 \\
Qwen3VL-4B & aro & 24 & 3,560 & 0.0665 & -0.0161 & [-0.0165, -0.0158] & 5.7 \\
Qwen3VL-4B & ARO-relation slice & 24 & 1,517 & 0.0702 & -0.0119 & [-0.0125, -0.0113] & 13.4 \\
Qwen3VL-4B & sugarcrepe & 24 & 3,572 & 0.0693 & -0.0151 & [-0.0155, -0.0147] & 7.2 \\
Qwen3VL-4B & winoground & 24 & 577 & 0.0651 & -0.0106 & [-0.0117, -0.0094] & 22.2 \\
Qwen3VL-8B & aro & 32 & 3,649 & 0.1209 & 0.0333 & [0.0322, 0.0344] & 88.1 \\
Qwen3VL-8B & ARO-relation slice & 32 & 1,602 & 0.1064 & 0.0226 & [0.0208, 0.0243] & 75.5 \\
Qwen3VL-8B & sugarcrepe & 32 & 3,839 & 0.1466 & 0.0330 & [0.0321, 0.0338] & 92.5 \\
Qwen3VL-8B & winoground & 32 & 579 & 0.1314 & 0.0011 & [-0.0015, 0.0036] & 47.0 \\
Gemma3-4B & aro & 27 & 3,133 & 0.0095 & -0.0012 & [-0.0013, -0.0012] & 17.8 \\
Gemma3-4B & ARO-relation slice & 27 & 1,420 & 0.0095 & -0.0013 & [-0.0014, -0.0012] & 17.5 \\
Gemma3-4B & sugarcrepe & 27 & 3,965 & 0.0090 & -0.0020 & [-0.0021, -0.0020] & 5.5 \\
Gemma3-4B & winoground & 27 & 550 & 0.0090 & -0.0008 & [-0.0010, -0.0007] & 24.9 \\
InternVL3-2B & aro & 23 & 3,484 & 0.0358 & -0.0419 & [-0.0427, -0.0412] & 3.8 \\
InternVL3-2B & ARO-relation slice & 23 & 1,755 & 0.0411 & -0.0355 & [-0.0366, -0.0344] & 5.9 \\
InternVL3-2B & sugarcrepe & 23 & 3,974 & 0.0382 & -0.0422 & [-0.0428, -0.0416] & 0.5 \\
InternVL3-2B & winoground & 23 & 572 & 0.0410 & -0.0247 & [-0.0264, -0.0229] & 9.3 \\
\bottomrule
\end{tabular}
\caption{Source/slice-level paired excess breakdown. Winoground, SugarCrepe, and ARO are the public source pools; ARO-relation is an analysis slice within ARO, not an additional public source. The table is intended for diagnostic interpretation rather than source leaderboarding.}
\label{tab:app_paired_source}
\end{table*}

\subsection{Stress-Type Breakdown}
Table~\ref{tab:app_paired_stress_type} reports the same diagnostic by stress type. The main purpose is to assess whether the paired-excess signal is specific to a single perturbation category or broadly present across object, attribute, relation, and compositional stress. Because some compositional slices are small, these rows are retained for transparency rather than used for strong standalone claims.

\begin{table*}[h]
\centering
\tiny
\begin{tabular}{llrrrrrr}
\toprule
\textbf{Model} & \textbf{Stress type} & \textbf{Layer} & \textbf{Strict N} & \textbf{\dlex{} mean} & \textbf{\drand{} mean} & \textbf{\drand{} 95\% CI} & \textbf{$\%>0$} \\
\midrule
Qwen3VL-2B & attribute & 24 & 3,212 & 0.0191 & -0.0162 & [-0.0165, -0.0159] & 2.7 \\
Qwen3VL-2B & compositional & 24 & 75 & 0.0187 & -0.0184 & [-0.0208, -0.0159] & 5.3 \\
Qwen3VL-2B & object & 24 & 3,088 & 0.0201 & -0.0156 & [-0.0160, -0.0153] & 3.5 \\
Qwen3VL-2B & relation & 24 & 3,286 & 0.0180 & -0.0149 & [-0.0153, -0.0146] & 9.6 \\
Qwen3VL-4B & attribute & 24 & 3,232 & 0.0674 & -0.0160 & [-0.0164, -0.0155] & 6.2 \\
Qwen3VL-4B & compositional & 24 & 69 & 0.0683 & -0.0140 & [-0.0164, -0.0116] & 7.2 \\
Qwen3VL-4B & object & 24 & 3,054 & 0.0683 & -0.0151 & [-0.0155, -0.0147] & 6.6 \\
Qwen3VL-4B & relation & 24 & 2,871 & 0.0687 & -0.0128 & [-0.0133, -0.0124] & 13.4 \\
Qwen3VL-8B & attribute & 32 & 3,315 & 0.1332 & 0.0345 & [0.0335, 0.0355] & 91.7 \\
Qwen3VL-8B & compositional & 32 & 87 & 0.1377 & 0.0163 & [0.0111, 0.0214] & 73.6 \\
Qwen3VL-8B & object & 32 & 3,134 & 0.1288 & 0.0299 & [0.0289, 0.0309] & 89.2 \\
Qwen3VL-8B & relation & 32 & 3,133 & 0.1255 & 0.0241 & [0.0228, 0.0254] & 75.0 \\
Gemma3-4B & attribute & 27 & 3,128 & 0.0093 & -0.0016 & [-0.0016, -0.0015] & 12.7 \\
Gemma3-4B & compositional & 27 & 80 & 0.0089 & -0.0016 & [-0.0021, -0.0013] & 10.0 \\
Gemma3-4B & object & 27 & 2,786 & 0.0095 & -0.0014 & [-0.0015, -0.0014] & 13.0 \\
Gemma3-4B & relation & 27 & 3,074 & 0.0091 & -0.0017 & [-0.0017, -0.0016] & 12.9 \\
InternVL3-2B & attribute & 23 & 3,191 & 0.0339 & -0.0452 & [-0.0459, -0.0445] & 1.3 \\
InternVL3-2B & compositional & 23 & 65 & 0.0378 & -0.0318 & [-0.0369, -0.0263] & 7.7 \\
InternVL3-2B & object & 23 & 3,122 & 0.0416 & -0.0373 & [-0.0380, -0.0365] & 3.3 \\
InternVL3-2B & relation & 23 & 3,407 & 0.0387 & -0.0374 & [-0.0382, -0.0367] & 4.6 \\
\bottomrule
\end{tabular}
\caption{Stress-type paired excess breakdown. Sparse slices are excluded from strong standalone claims and retained for transparency.}
\label{tab:app_paired_stress_type}
\end{table*}

\section{Hidden Stress-Sensitivity Thresholding}
\label{app:hsr_thresholds}

HSR depends on a threshold $\tau$, so we interpret it as a summary statistic rather than the primary statistical test. Percentile rules such as $p80$, $p90$, and $p95$ are distributional sensitivity checks; by construction they yield approximately $20\%$, $10\%$, and $5\%$ high-score subsets. The non-percentile mean-plus-standard-deviation rule is shown below for the completed high-support models.
\begin{equation}
\mathrm{HSR}_{\mathrm{correct}}(\tau)=\Pr(S(h)>\tau \mid \mathrm{strict\text{-}correct}=1).
\end{equation}

\begin{table*}[h]
\centering
\small
\begin{tabular}{llrrrr}
\toprule
\textbf{Model} & \textbf{Score} & \textbf{$\tau$} & \textbf{High / N} & \textbf{HSR (\%)} & \textbf{Mean high} \\
\midrule
Qwen3VL-2B & \dlex{} & 0.0280 & 1624 / 9661 & 16.8 & 0.034 \\
Qwen3VL-2B & \drand{} & -0.0059 & 1588 / 9661 & 16.4 & -0.001 \\
Qwen3VL-2B & Order instability & 0.0480 & 1795 / 9661 & 18.6 & 0.057 \\
Qwen3VL-4B & \dlex{} & 0.0840 & 1477 / 9226 & 16.0 & 0.092 \\
Qwen3VL-4B & \drand{} & -0.0030 & 1349 / 9226 & 14.6 & 0.002 \\
Qwen3VL-4B & Order instability & 0.1396 & 1483 / 9226 & 16.1 & 0.156 \\
Qwen3VL-8B & \dlex{} & 0.1748 & 1440 / 9669 & 14.9 & 0.198 \\
Qwen3VL-8B & \drand{} & 0.0613 & 1428 / 9669 & 14.8 & 0.080 \\
Qwen3VL-8B & Order instability & 0.0740 & 1767 / 9669 & 18.3 & 0.086 \\
Gemma3-4B & \dlex{} & 0.0110 & 1226 / 9068 & 13.5 & 0.012 \\
Gemma3-4B & \drand{} & 0.0001 & 1044 / 9068 & 11.5 & 0.001 \\
Gemma3-4B & Order instability & 0.0124 & 1151 / 9068 & 12.7 & 0.013 \\
InternVL3-2B & \dlex{} & 0.0570 & 1872 / 9785 & 19.1 & 0.067 \\
InternVL3-2B & \drand{} & -0.0183 & 1733 / 9785 & 17.7 & -0.007 \\
InternVL3-2B & Order instability & 0.1332 & 2154 / 9785 & 22.0 & 0.157 \\
\bottomrule
\end{tabular}
\caption{Mean-plus-standard-deviation thresholding for selected hidden stress-sensitivity summaries under stress-trial strict-correct behavior. HSR is used as a compact exposure summary; paired distributional contrasts remain the primary evidence.}
\label{tab:app_hsr_mean_std}
\end{table*}
\section{Perturbation Sanity-Check Details}
\label{app:perturbation_details}

The perturbation evidence is used as a preliminary sanity check, not as a complete mechanism claim. A stress-sensitive direction is compared against a random-control direction or another random-control baseline. The reported quantity is the excess margin perturbation:
\begin{equation}
\Delta^{\mathrm{int}} = \mathrm{Effect}(v_{\mathrm{stress}})-\mathrm{Effect}(v_{\mathrm{random}}).
\end{equation}
Here \emph{Effect} is the mean absolute change in expected answer margin under perturbation. Available runs use strict-correct examples, alpha values $\{0,0.25,0.5,1.0\}$, seed 42, one seeded random direction, and PC1-derived stress/preserve/lexical directions. We therefore report these as limited perturbation sanity checks rather than as evidence of a general decision mechanism.

\begin{table*}[h]
\centering
\small
\begin{tabular}{lrrrrrrr}
\toprule
\textbf{Model} & \textbf{Layer} & \textbf{Pairs} & \textbf{Excess} & \textbf{95\% CI} & \textbf{$>$0 (\%)} & \textbf{Stress flip (\%)} & \textbf{Rand. flip (\%)} \\
\midrule
Qwen3VL-4B & 24 & 1000 & 1.22 & [1.15, 1.28] & 84.8 & 3.3 & 0.1 \\
Qwen3VL-4B & 26 & 400 & 8.72 & [8.15, 9.31] & 99.8 & 47.0 & 0.0 \\
Qwen3VL-4B & 27 & 400 & 9.34 & [8.68, 10.03] & 99.2 & 48.5 & 0.0 \\
Qwen3VL-4B & 28 & 400 & 8.61 & [8.07, 9.17] & 100.0 & 47.0 & 0.0 \\
Gemma3-4B & 30 & 400 & 14.25 & [12.85, 15.68] & 75.8 & 48.0 & 0.2 \\
Gemma3-4B & 27 & 400 & 15.90 & [14.61, 17.22] & 100.0 & 48.0 & 0.5 \\
\bottomrule
\end{tabular}
\caption{Bootstrap confidence intervals for E5 stress-direction versus random-direction perturbation effects. The sharp excess effects near Qwen3VL-4B layers 26--28 and Gemma3-4B layer 27 are consistent with the limited sanity-check claim that one sampled random-control direction does not explain the observed margin perturbation; they do not establish a general causal mechanism or decision mechanism.}
\label{tab:app_e5_bootstrap}
\end{table*}
\section{Raw Evidence for \CA{}/\SO{}/\SI{}/\BE{} Profiles}
\label{app:four_dim_raw}

Table~\ref{tab:four_dim_raw} reports the raw evidence for \CA{}/\SO{}/\SI{}/\BE{} profiles.

\begin{table*}[h]
\centering
\small
\begin{tabular}{lrrrrr}
\toprule
\textbf{Metric} & \textbf{Qwen3VL-2B} & \textbf{Qwen3VL-4B} & \textbf{Qwen3VL-8B} & \textbf{Gemma3-4B} & \textbf{InternVL3-2B} \\
\midrule
E2B PC1                    & 0.167  & 0.378  & 0.252  & 0.204  & 0.369  \\
Rank drop                  & 10.06  & 16.51  & 19.62  & 6.67   & 14.02  \\
Best layer                 & 24     & 31     & 32     & 27     & 23     \\
E3 stress$-$pres PC1 gap   & 0.167  & 0.301  & 0.198  & 0.204  & 0.369  \\
E3 stress$-$lex PC1 gap    & 0.249  & 0.227  & $-$0.239 & 0.271 & 0.383  \\
Peak layer                 & 24     & 24     & 32     & 27     & 23     \\
E4 split-half overlap      & 1.000  & 1.000  & 1.000  & 1.000  & 1.000  \\
E4 stress$-$pres dir overlap & 0.194 & 0.069 & 0.301 & 0.264 & 0.231 \\
E4 stress$-$lex dir overlap  & 0.193 & 0.040 & 0.228 & 0.417 & 0.035 \\
E5 perturbation excess     & --     & 9.34   & --     & 15.90  & --     \\
\bottomrule
\end{tabular}
\caption{Raw evidence fields used to construct the \CA{}/\SO{}/\SI{}/\BE{} diagnostic profile. E2B summarizes \CA{} evidence; E3 and E4 summarize \SI{} and \SO{} evidence; E5 reports intervention sanity-check evidence in the completed runs. The E3 PC1 gap rows report condition-level differences in PC1 variance concentration (e.g., PC1(stress)~$-$~PC1(lexical)) and serve as profile-level selectivity diagnostics; they are \emph{not} the paired preserve-anchored $\Delta^{\text{lex}}$ reported in Table~\ref{tab:paired_excess_main} and Eq.~\ref{eq:dlex}, and should not be compared directly in sign or magnitude with that quantity. \SI{} is normalized from these PC1 gaps and the E4 direction-overlap terms; it is not computed from the paired $\Delta^{\text{lex}}$ contrast. E4 split-half overlap is 1.000 across all reported runs and is omitted from the table. Incomplete or low-support runs are omitted from the main evidence.}\label{tab:four_dim_raw}
\end{table*}

\section{Model and Implementation Details}
\label{app:implementation}

\subsection{Model Checkpoints and Compute Budget}
\label{app:model_compute}

All main evidence is based on white-box multimodal language models for which pre-answer hidden states are accessible. We perform inference-only evaluation: no model training, fine-tuning, gradient-based optimization, or hyperparameter search is conducted. Generation is disabled for the forced-choice decision. For each prompt, we use the forward-pass logits over the A/B answer tokens to determine the model's choice, and we extract hidden states immediately before the answer token.

Each source record contains a positive reference caption and four candidate conditions: preserve, stress, lexical, and random. Each record--candidate pair is evaluated under both original and swapped option orders. Thus, for each model, the completed high-support run contains $11{,}522 \times 4 = 46{,}088$ record--condition pairs and $46{,}088 \times 2 = 92{,}176$ A/B forced-choice forward passes. Table~\ref{tab:app_checkpoint_metadata} reports the checkpoint and implementation metadata, and Table~\ref{tab:app_compute_budget} reports the approximate compute budget.

\begin{table*}[h]
\centering
\scriptsize
\begin{tabular}{llp{5.2cm}lll}
\toprule
\textbf{Model} & \textbf{Param. scale} & \textbf{Checkpoint} & \textbf{dtype} & \textbf{Max new} & \textbf{Representation site} \\
\midrule
Qwen3VL-2B & $\sim$2B & \texttt{Qwen/Qwen3-VL-2B-Instruct} & bfloat16 & 4 & pre\_answer\_token \\
Qwen3VL-4B & $\sim$4B & \texttt{Qwen/Qwen3-VL-4B-Instruct} & bfloat16 & 4 & pre\_answer\_token \\
Qwen3VL-8B & $\sim$8B & \texttt{Qwen/Qwen3-VL-8B-Instruct} & bfloat16 & 4 & pre\_answer\_token \\
Gemma3-4B & $\sim$4B & \texttt{google/gemma-3-4b-it} & bfloat16 & 4 & pre\_answer\_token \\
InternVL3-2B & $\sim$2B & \texttt{OpenGVLab/InternVL3-2B-hf} & bfloat16 & 4 & pre\_answer\_token \\
\bottomrule
\end{tabular}
\caption{Checkpoint and implementation metadata for the completed high-support runs. Parameter scale follows the nominal model size indicated by the released checkpoint name. All listed runs use bfloat16 inference, the model-specific default processor, forward-pass A/B logits without sampling, and hidden-state extraction at the pre-answer-token representation site. Exact Hugging Face revision SHA values were not consistently stored in the local artifacts; checkpoint identifiers and configuration hashes are available in run metadata.}
\label{tab:app_checkpoint_metadata}
\end{table*}

\begin{table*}[h]
\centering
\scriptsize
\begin{tabular}{lrrrr}
\toprule
\textbf{Model} & \textbf{Records} & \textbf{Cond. pairs} & \textbf{A/B forwards} & \textbf{Est. GPU-hours} \\
\midrule
Qwen3VL-2B & 11,522 & 46,088 & 92,176 & $\sim$20 \\
Qwen3VL-4B & 11,522 & 46,088 & 92,176 & $\sim$20 \\
Qwen3VL-8B & 11,522 & 46,088 & 92,176 & $\sim$20 \\
Gemma3-4B & 11,522 & 46,088 & 92,176 & $\sim$20 \\
InternVL3-2B & 11,522 & 46,088 & 92,176 & $\sim$30 \\
\midrule
\textbf{Total} & 57,610 & 230,440 & 460,880 & $\sim$110 \\
\bottomrule
\end{tabular}
\caption{Approximate compute budget for the completed high-support runs. Each run used one NVIDIA RTX A6000 GPU, so estimated wall-clock hours and estimated GPU-hours are numerically the same. The reported budget covers inference-time forced-choice evaluation and hidden-state extraction. It does not include model training, fine-tuning, hyperparameter search, or downstream analysis scripts whose cost is negligible relative to model forward passes. The runtimes are approximate because exact scheduler logs were not preserved for all runs.}
\label{tab:app_compute_budget}
\end{table*}

Image preprocessing follows the default processor associated with each model checkpoint unless otherwise specified by the run configuration. The local artifacts store checkpoint identifiers, configuration hashes, and a common code commit, but not all server-side job metadata such as exact scheduler records or revision-level Hugging Face SHA values.

Unless otherwise stated, each model result corresponds to one deterministic inference pass over the fixed evaluation subset. Reported confidence intervals quantify item-level uncertainty through paired bootstrap resampling over source records after averaging original and swapped option orders; they do not measure variance across independently trained models, random seeds, or checkpoint revisions.

\subsection{Experimental Setup and Hyperparameters}
\label{app:setup_hyperparameters}

The experiments are inference-only. No model training, fine-tuning, gradient-based optimization, or hyperparameter search is performed, so there are no optimizer, learning-rate, epoch, or weight-decay settings. The relevant experimental settings are the fixed evaluation subset, prompt format, answer-extraction rule, representation site, layer-selection rule, and bootstrap procedure.

Each source record is evaluated under four candidate conditions---preserve, stress, lexical, and random---and each record--condition pair is evaluated in both original and swapped A/B option orders. The prompt asks the model to answer only A or B. We use deterministic forward-pass logits over the A/B answer tokens for the forced-choice decision and extract hidden states at the pre-answer-token representation site. All models use bfloat16 inference on a single NVIDIA RTX A6000 GPU with the default processor and image preprocessing associated with each checkpoint. We set \texttt{max\_new\_tokens=4} only for answer extraction; no temperature, top-$p$, top-$k$, beam search, or sampling setting is tuned.

For selected-layer reporting, the diagnostic layer is chosen once per model by the fixed stress-specificity localization rule and then frozen for all selected-layer contrasts. Cosine distances are computed directly on pre-answer hidden states without centering, whitening, dimensionality reduction, learned probes, or generated-text pooling. Paired excess contrasts are computed after averaging original and swapped orders within each source record. Confidence intervals use paired bootstrap resampling over source-record units with 3,000 resamples. Unless otherwise stated, each model result is one deterministic inference pass over the fixed subset; intervals quantify item-level uncertainty, not variance across training seeds or checkpoint revisions.

\subsection{Software packages and package parameters.}
All experiments were implemented in Python using PyTorch, Hugging Face Transformers, and model-specific wrappers where required for model loading, multimodal preprocessing, tokenization, forward passes, answer extraction, and hidden-state collection. For each model, we used the model-specific processor/tokenizer and chat template when available, rather than applying a separate caption-processing pipeline. Image resizing, normalization, multimodal prompt formatting, and vision-token construction follow the default processor associated with each released checkpoint unless otherwise specified by the run configuration. Model outputs were decoded only to parse the forced-choice A/B answer when needed; all latent analyses used hidden states collected from the pre-answer decision state.

We used NumPy and Pandas for array operations, table construction, aggregation, and CSV/JSONL processing. Confidence intervals were computed with nonparametric paired bootstrap resampling using our analysis scripts. Unless otherwise specified, bootstrap intervals use 3,000 resamples, and the source item after original/swapped order averaging, rather than the layer or individual A/B forward pass, is the resampling unit. Cosine distances, paired excess scores, and, where reported, PC1 variance ratios and effective-rank summaries were computed directly from extracted hidden-state arrays using standard tensor or NumPy linear-algebra routines. We do not center, whiten, train probes, or apply dimensionality reduction before computing the primary paired distance scores.

We did not use NLP evaluation packages such as NLTK, spaCy, ROUGE, BLEU, METEOR, CIDEr, BERTScore, or learned caption scorers to compute the reported S$^3$E metrics. The lexical-control and surface-control analyses rely only on the benchmark-provided captions and lightweight token-overlap statistics computed by our scripts. Thus, the reported decision-state stress-sensitivity measures are not derived from external text-similarity, caption-quality, or free-form generation metrics. Package versions are specified in the released environment file.

\end{document}